%% file: main.tex
\documentclass[10pt,twocolumn,letterpaper]{article}

\usepackage[pagenumbers]{iccv}

\input{preamble}
\input{macros}

\definecolor{iccvblue}{rgb}{0.21,0.49,0.74}
\usepackage[pagebackref,breaklinks,colorlinks,allcolors=iccvblue]{hyperref}

\title{\paperTitle}

\author{\authorBlock}

\begin{document}
\maketitle

\begin{strip}
    \centering
    \vspace{-35pt}
    \includegraphics[width=\linewidth, trim={0 0 0 0},clip]{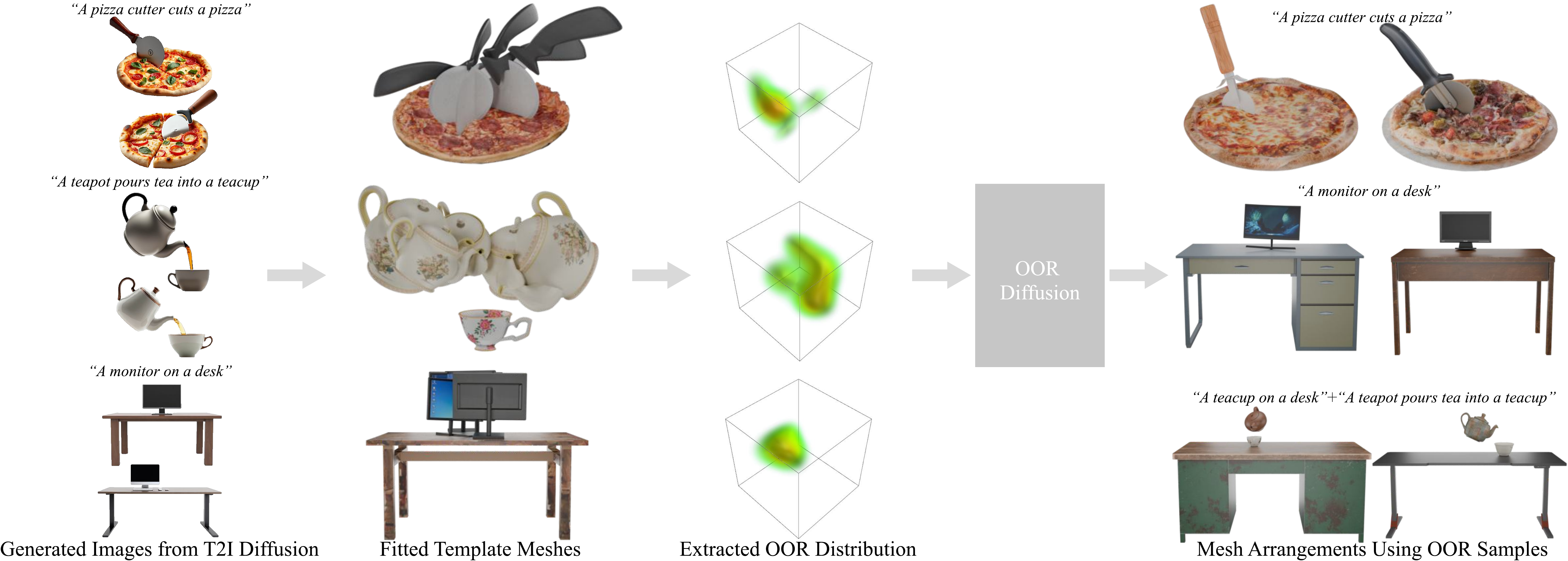}
    \captionof{figure}{\textbf{Object-Object Spatial Relationships (OOR).} Given a textual description of the spatial relationship between two objects, our method models OOR, representing their relative poses and scales with respect to the text. We obtain synthetic OOR samples using off-the-shelf models and a proposed mesh registration method, then learn their distribution through OOR diffusion. During inference, our OOR diffusion model generates OOR samples conditioned on the text input.}
    \label{fig:teaser}
    \vspace{-10pt}
\end{strip}

\input{sec/0_abstract}    
\input{sec/1_intro}
\input{sec/2_related}
\input{sec/3_method}
\input{sec/4_experiments}
\input{sec/5_conclusion}

\section*{Acknowledgements}
We thank Inhee Lee and Taeksoo Kim for their assistance in conducting the user study. This work was supported by RLWRLD, NRF grant funded by the Korean government (MSIT) (No. RS-2022-NR070498 and RS-2025-25396144), and IITP grant funded by the Korea government (MSIT) (No. RS-2024-00439854, No.RS-2021-II211343, No.RS-2025-25441838, and No.2022-0-00156). H. Joo is the corresponding author.

{
    \small
    \bibliographystyle{ieeenat_fullname}
    \bibliography{main}
}

\input{sec/10_supp}

\end{document}

%% file: macros.tex
\usepackage{times}
\usepackage{microtype}
\usepackage{epsfig}
\usepackage[skip=5pt]{caption}
\usepackage{float}
\usepackage{placeins}
\usepackage{color, colortbl}
\usepackage{stfloats}
\usepackage{enumitem}
\usepackage{tabularx}
\usepackage{xstring}
\usepackage{cuted} %
\usepackage{multirow}
\usepackage{xspace}
\usepackage{url}
\usepackage{mathtools}
\usepackage[hang,flushmargin]{footmisc}
\usepackage[normalem]{ulem}  %
\usepackage{graphicx}
\usepackage{amsmath}
\usepackage{amssymb}
\usepackage{comment}
\usepackage[title]{appendix}
\usepackage{cancel}
\usepackage{bbm}
\usepackage{afterpage}
\usepackage{algorithm}
\usepackage{algorithmic}
\usepackage{bm}
\usepackage{gensymb}

\newfloat{sepfigure}{p}{lop}
\floatname{sepfigure}{Figure}

\definecolor{R1color}{rgb}{0,0.56,0}
\newcommand{\R}[1]{{%
    \textbf{%
        \ifstrequal{#1}{1}{\textcolor[rgb]{0,0.56,0}{R#1}}{%
        \ifstrequal{#1}{2}{\textcolor{blue}{R#1}}{%
        \ifstrequal{#1}{3}{\textcolor{magenta}{R#1}}{%
        \ifstrequal{#1}{4}{\textcolor{teal}{R#1}}{%
                           \textcolor{cyan}{R#1}%
        }}}}%
    }%
}}

\usepackage{algorithm}
\usepackage{algorithmic}

\usepackage{bm}

\def\paperID{3511} %
\def\confName{ICCV}
\def\confYear{2025}

\def\paperTitle{Learning 3D Object Spatial Relationships from Pre-trained 2D Diffusion Models}

\def\authorBlock{
    Sangwon Baik$^{1}$ \qquad
    Hyeonwoo Kim$^{1}$ \qquad
    Hanbyul Joo$^{1,2}$ \\
    [0.4em] %
    $^1$Seoul National University \quad $^2$RLWRLD \\
    [0.25em] %
    {\tt\small \href{https://tlb-miss.github.io/oor/}{\color{magenta}{https://tlb-miss.github.io/oor/}}}
}

%% file: sec/0_abstract.tex
\begin{abstract}
We present a method for learning 3D spatial relationships between object pairs, referred to as object-object spatial relationships (OOR), by leveraging synthetically generated 3D samples from pre-trained 2D diffusion models. We hypothesize that images synthesized by 2D diffusion models inherently capture realistic OOR cues, enabling efficient collection of a 3D dataset to learn OOR for various unbounded object categories. Our approach synthesizes diverse images that capture plausible OOR cues, which we then uplift into 3D samples. Leveraging our diverse collection of 3D samples for the object pairs, we train a score-based OOR diffusion model to learn the distribution of their relative spatial relationships. Additionally, we extend our pairwise OOR to multi-object OOR by enforcing consistency across pairwise relations and preventing object collisions. Extensive experiments demonstrate the robustness of our method across various object-object spatial relationships, along with its applicability to 3D scene arrangement tasks and human motion synthesis using our OOR diffusion model.
\end{abstract}

%% file: sec/1_intro.tex
\section{Introduction}
\label{sec:intro}

In real-world scenes, specific spatial and functional patterns exist between objects. While some are constrained by physical laws (e.g., objects can rest on others but cannot float in mid-air), many arise from functional usage, reflecting how humans interact with and arrange these objects. For instance, chairs are commonly positioned around tables, and items such as cups and bottles are usually placed on tables rather than chairs, despite the physical possibility of alternative arrangements. More sophisticated patterns also exist; for example, a pizza cutter is typically angled relative to the pizza to facilitate slicing across its diameter (as shown in Fig.~\ref{fig:teaser}). These intuitive yet diverse relationships, which we refer to as object-object spatial relationships (OOR), capture the relative poses and scales between object pairs. Enabling machines to understand and generate these natural layouts and spatial relationships is crucial for various applications such as content creation, immersive VR/AR, and robotic manipulation tasks. However, the diversity of object spatial relationships, across object categories, contexts, and scenarios, makes them challenging to model through manual annotation or data collection in controlled setups.

One promising approach is to learn spatial relationships between objects from 2D images. However, raw images from the Internet are often cluttered and overly ``wild'', complicating the learning of precise 3D spatial relationships between objects. To overcome this limitation, we present an approach to learn 3D object spatial relationships from synthetically generated 3D samples capturing plausible OORs. 
Our method is inspired by recent approaches that pursue human-object interaction and affordances through synthetic images generated by pre-trained image diffusion models~\cite{ComA, CHORUS}, where the 2D diffusion model offers effective ways to generate samples for learning such relations.

We first rigorously formalize OOR as the relative scale and pose between two object categories.
To address the scarcity of 3D data for learning OORs across diverse object pairs, we introduce a framework that generates diverse and realistic 3D spatial relationship samples from synthesized 2D images, eliminating the need for labor-intensive data collection or manual annotation. Our 3D lifting pipeline robustly reconstructs 3D samples from synthetically generated 2D images of object-pairs.
Next, we present a text-conditioned OOR diffusion based on a score-based model~\cite{song2021scorebased, zhang2024generative} in the spatial parametric space, capable of modeling relative rotations, translations, and scales between objects across diverse scenarios.
Our OOR diffusion generates varied OOR patterns conditioned on a text prompt describing the scene context.
To enhance generalization, we leverage LLM~\cite{openai2024chatgpt} for text context augmentation, expanding our dataset to 475 distinct OOR scenarios. In addition, we extend our pairwise OOR modeling to multi-object settings, capturing spatial relationships among multiple objects in a shared 3D space. Through extensive experiments, we demonstrate the robustness of our method across various object-object spatial relationships. We also present several 3D scene arrangement tasks and the synthesis of human motion interacting with two objects as applications of our OOR diffusion model.

In summary, our main contributions are as follows: (1) We formulate a novel representation for object-object spatial relationships (OOR); (2) We introduce an effective pipeline to generate diverse 3D OOR data from synthetic images, incorporating prompting strategies, data augmentation, and filtering; (3) We propose a text-conditioned score-based diffusion model to effectively model the OOR distribution. To improve generalization across diverse OOR scenarios, we incorporate LLM-based text augmentation. We also extend pairwise OOR modeling to the multi-object setting via an optimization strategy guided by novel inference-time losses; (4) We demonstrate the capability of our OOR diffusion model for 3D scene editing through optimization that directly leverages the score function of the underlying distribution. We further show that the generated OOR can be used to synthesize human motions interacting with two objects.

%% file: sec/2_related.tex
\section{Related Work}
\label{sec:rel_work}

\noindent \textbf{Learning Object Spatial Relationships.}
In robotics, research on spatial relations between objects has primarily focused on teaching robots to learn and replicate object placement~\cite{rosman2011learning, kartmann2021semantic, pek2023spatial, luo2023grounding, mees2020learning, hou2024lop, kiran2022spatial, wen2024can, NEURIPS2022_819aaee1}. However, these methods often overlook complex relationships, like "a teapot pouring tea into a teacup." In object detection, there is a growing emphasis on leveraging spatial relationships between objects to improve detection performance, particularly for identifying occluded or challenging-to-recognize objects~\cite{zhao2023rgrn, li2020gar}. For indoor scenes, the availability of 3D datasets~\cite{dai2017scannet, song2015sun, fu20213d, roberts2021hypersim, replica19arxiv, hua2016scenenn} capturing object-object spatial relationships has accelerated progress in layout estimation~\cite{zou2018layoutnet, lee2017roomnet}, CAD model retrieval~\cite{gao2024diffcad, gumeli2022roca, maninis2022vid2cad, langer2022sparc, izadinia2017im2cad, kim2012acquiring}, and scene generation~\cite{paschalidou2021atiss, yang2021scene}. However, these methods are limited to predefined object categories, whereas our approach addresses unbounded object pairs.

\noindent \textbf{Learning from Pre-trained 2D Diffusion Models.}
The advent of diffusion models~\cite{sohl2015deep, ho2020denoising, dhariwal2021diffusion} has significantly boosted performance and accelerated progress in the field. Accordingly, many studies have focused on leveraging the rich 2D knowledge in pre-trained diffusion models to extract insights and enable novel applications.
CHORUS~\cite{CHORUS} and ComA~\cite{ComA} propose learning human-object relationships and interactions from inconsistent 2D image pools generated by diffusion. However, unlike humans, estimating object poses is more difficult, making our task more challenging.
Additionally, various approaches have been developed to extract semantic features from images using pre-trained diffusion models~\cite{NEURIPS2022_8aff4ffc, Xiang_2023_ICCV, zhang2023tale, zhang2024telling, mariotti2024improving, Cha_2025_CVPR}.

\noindent \textbf{Score-based Diffusion Models.}
A score-based generative model~\cite{song2019generative, song2020improved} models a distribution using the score function, often more tractable than computing the likelihood directly. It was integrated with diffusion models in~\cite{song2021scorebased}. Score-based methods are widely adopted not only for image generation but also for pose generation and estimation of objects~\cite{zhang2024generative, Hsiao_2024_CVPR} and humans~\cite{Shan_2023_ICCV, dposer, Ci_2023_CVPR}.
In particular, GenPose~\cite{zhang2024generative}, which estimates the 6D pose of a point cloud, serves as the backbone of our OOR diffusion. DAViD~\cite{david} applies score-based modeling to the HOI domain by leveraging GenPose.

%% file: sec/3_method.tex
\section{Method}
\label{sec:method}
We present a method for learning object-object spatial relationships (OOR) from synthetically generated 2D images by pre-trained diffusion models. We model OOR based on relative poses and scales of a pair of objects in canonical space. In Sec.~\ref{sec:formulation} we formalize our definition of OOR. Then, in Sec.~\ref{sec:dbgen}, we describe our pipeline for constructing 3D OOR dataset by generating 2D synthetic images and uplifting them to 3D. Finally, in Sec.~\ref{sec:diffusion}, we present our OOR diffusion model trained on the generated 3D OOR dataset to learn the distribution of object-object relationships. We further extend our approach from pairwise to multi-object OORs, enabling the modeling of spatial relationships among multiple objects within a shared space.

\subsection{Formulating Object-Object Relationship}
\label{sec:formulation}
\begin{figure}[t]
    \centering
    \includegraphics[width=\linewidth, trim={0 0 0 0},clip]{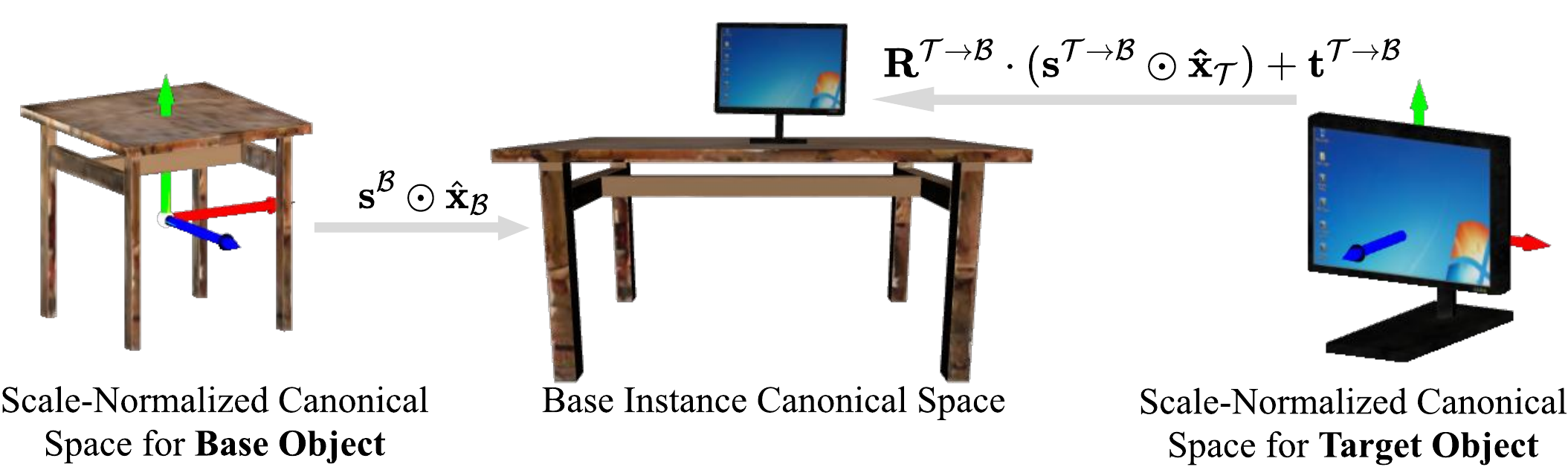}
    \captionof{figure}{\textbf{
    Our Coordinate Systems.} We conceptually model the transformation from a scale-normalized space, where the tightest 3D bounding box of each object is normalized to the unit cube, to a canonical space for each object instance within an object category.
    }
    \label{fig:Space Conversion}
    \vspace{-15pt}
\end{figure}

We define OOR as the relative poses and scales between a pair of object categories specified by a text prompt $\mathbf{c}$ describing object categories and their spatial relationships. 
Each 3D object instance is placed in an \textit{instance canonical space} whose origin is the midpoint of its tightest 3D bounding box.
The 3D Bbox's up-vector aligns with the $y$-axis, and its ground plane is parallel with the $x$-$z$ plane.
The frontal side, typically the most observable view, faces the $z$-axis, although our method accommodates any canonical orientation. The scale is normalized such that the longest edge of the 3D Bbox is set to $1$, while keeping the aspect ratios of the objects. 
See Fig.~\ref{fig:Space Conversion} for an example.
To account for varying aspect ratios within a category, we further consider a \textit{scale-normalized canonical space} that normalizes the 3D Bbox to a unit cube.
We consider the per-instance scaling factor $\mathbf{s} \in \mathbb{R}^3_+$ which recover the original aspect ratio from the scale-normalized one: $\mathbf{x}  = \mathbf{s}  \odot \mathbf{\hat{x}}$, where $\mathbf{\hat{x}} \in \mathbb{R}^3$ is the 3D point in the scale-normalized canonical space, $\mathbf{x} \in \mathbb{R}^3$ is the 3D point in the original instance canonical space with original aspect ratio, and $\odot$ denotes the element-wise product. 

Given a pair of objects, we denote one as a base object and the other as a target, formulating OOR as the relative poses and scales of a target object with respect to a base object.
For example, given a context addressed as a text prompt $\mathbf{c}$, ``A teapot pouring tea into a teacup", we model the OOR as the relative poses and scales of the teapot (a target object) with respect to the teacup (a base object) in the teacup's canonical space. Note that we can consider either object as the base object without loss of generality. 

Specifically, given a text prompt $\mathbf{c}$ describing the OOR context, a base object category $\mathcal{B}$, and a target object category $\mathcal{T}$, their OOR distribution $p_\mathbf{c}^{\mathcal{T} \rightarrow \mathcal{B}}$ and OOR sample $\phi$ are defined as follows: 
\begin{gather}
    \mathbf{\phi} \sim p_\mathbf{c}^{\mathcal{T} \rightarrow \mathcal{B}}, \; \; \mathbf{\phi} = 
    (\mathbf{R}^{\mathcal{T} \rightarrow \mathcal{B}}, \mathbf{t}^{\mathcal{T} \rightarrow \mathcal{B}}, \mathbf{s}^{\mathcal{T} \rightarrow \mathcal{B}}, \mathbf{s}^{\mathcal{B}}),
    \label{eq:oor_formulation}
\end{gather}
where $\mathbf{R}^{\mathcal{T} \rightarrow \mathcal{B}} \in SO(3)$, $\mathbf{t}^{\mathcal{T} \rightarrow \mathcal{B}} \in \mathbb{R}^3$, and $\mathbf{s}^{\mathcal{T} \rightarrow \mathcal{B}} \in \mathbb{R}^3_+$ represent the rotation, translation, and scale transformations that convert a point $\mathbf{\hat{x}}_\mathcal{T} \in \mathcal{T}$ of the \emph{target object} (in the scale-normalized canonical space) into the instance canonical space of the \emph{base object}, resulting in $\mathbf{x}_{\mathcal{T} \rightarrow \mathcal{B}}$:
\begin{gather}\label{eqn:oor_target}
    \mathbf{x}_{\mathcal{T} \rightarrow \mathcal{B}} = 
    \mathbf{R}^{\mathcal{T} \rightarrow \mathcal{B}} \cdot (\mathbf{s}^{\mathcal{T} \rightarrow \mathcal{B}} \odot \mathbf{\hat{x}}_\mathcal{T}) + \mathbf{t}^{\mathcal{T} \rightarrow \mathcal{B}}.
\end{gather}
The $\mathbf{s}^{\mathcal{B}} \in \mathbb{R}^3_+$ is the scaling factor to convert from the point in the scale-normalized space, $\mathbf{\hat{x}}_\mathcal{B}$, into the instance space with the original aspect ratio of it:
\begin{gather}\label{eqn:oor_base}
    \mathbf{x}_\mathcal{B}  = \mathbf{s}^{\mathcal{B}}  \odot \mathbf{\hat{x}}_\mathcal{B}.
\end{gather}
Intuitively, an OOR sample $\mathbf{\phi}$ represents a plausible relative spatial relationship for the target object relative to the base object. The use of a non-isotropic scaling factor $\mathbf{s}^{\mathcal{B}} \in \mathbb{R}^3_+$ that accounts for the original aspect ratio of the 3D BBox is important, as OOR is inherently influenced by object 3D scale. For instance, placing items on a table requires knowing the dimensions of the tabletop and height, which we approximate with its 3D Bbox scale.

\begin{figure*}[t]
    \centering
    \includegraphics[width=\linewidth, trim={0 0 0 0},clip]{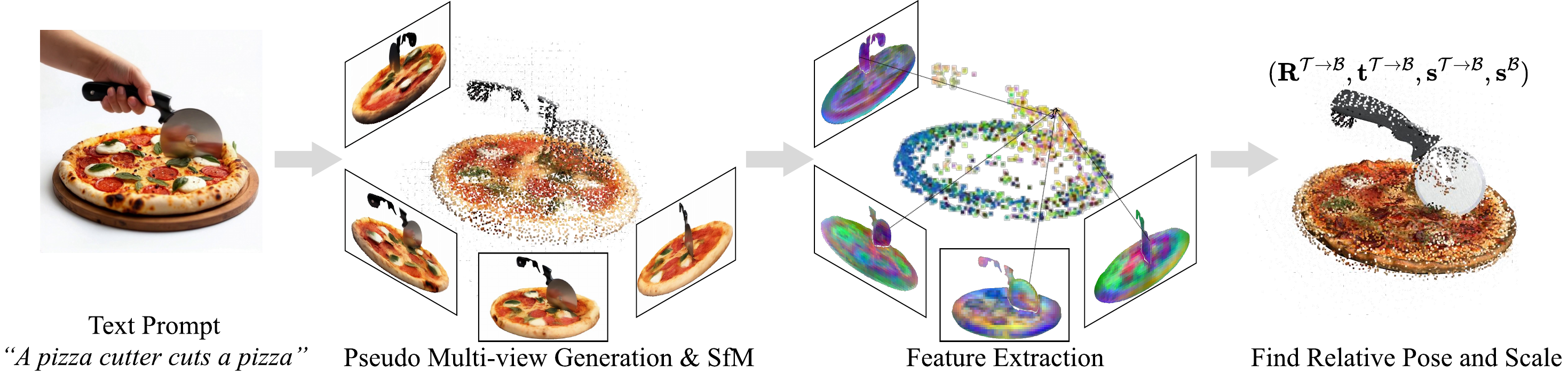}
    \captionof{figure}{\textbf{
    Dataset Generation Overview.} For a given text prompt related to an object pair, we obtain multi-view images and point clouds using off-the-shelf models. Then, we lift pixel features to obtain 3D point features. We repeat the same process for the rendered images of collected meshes. Finally, we perform Procrustes analysis with RANSAC to estimate the relative pose and scale of each object.
    }
    \label{fig:db_gen}
    \vspace{-15pt}
\end{figure*}

\subsection{3D OOR Samples Generation}
\label{sec:dbgen}

To effectively learn OOR defined in Eqs.~\ref{eq:oor_formulation}--~\ref{eqn:oor_base}, diverse 3D samples with plausible spatial relationships of each object pair are required. 
Obtaining such 3D samples from the real world is challenging.
We present a novel pipeline that synthesizes diverse 3D samples by leveraging pre-trained 2D diffusion models and an advanced 3D uplifting process.

\noindent \textbf{High-quality 2D OOR Images Synthesis.}
We use an off-the-shelf text-to-image model~\cite{flux} to generate images that are aligned to the OOR context in text prompt $\mathbf{c}$. We design the prompt with specific strategies to facilitate the later 3D lifting process, including: (1) appending ``white background" to the end of the prompt to ensure the full shape of each object is clearly visible; (2) incorporating object shape and texture descriptions to better align with our template object meshes; (3) Adjusting the viewpoint to mitigate image frame constraints for category pairs with large scale differences, such as (table, teacup). Additionally, we further diversify OORs by passing each image to an image-to-video model~\cite{KLING} and treating every frame as an additional 2D sample, ignoring temporal order. Further discussion of prompting and augmentation strategies is provided in Sec.~\ref{sec:supp_de_gen}

\noindent \textbf{Pseudo Multi-view Generation and SfM.} Given an image containing the OOR cues for the object pair, we produce pseudo-multi-view images using an off-the-shelf novel view synthesis method, SV3D~\cite{voleti2024sv3d}, which synthesizes circular multi-views from a single image input. These pseudo-multi-views look visually plausible often but may lack geometric consistency across views, so we reconstruct them with VGGSfM~\cite{wang2024vggsfm} and discard failures.
As the output of SfM, we obtain the 3D point cloud 
$\mathcal{P} = \{\mathbf{P}_j\}_{j=1}^N, \; \mathbf{P}_j \in \mathbb{R}^3$, and their corresponding 2D keypoints, $\{\mathbf{p}_j^k\}_{k=1}^{m_j}, \; \mathbf{p}_j^k \in \mathbb{R}^2$, where $N$ denotes the number of 3D points, and $m_j$ is the number of corresponding 2D keypoints for the $j$-th 3D point.

\noindent \textbf{Pose and Scale Extraction through Mesh Registration.} Given the point clouds $\mathcal{P}$ reconstructed in a certain coordinate system (denoted as point cloud space), we register the base object's template mesh $\mathcal{M}_{\mathcal{B}}$ and the target object's template mesh $\mathcal{M}_{\mathcal{T}}$ from their canonical spaces to the point cloud. From this registration, we compute their relative spatial transformation $T_{\text{rel}}$, which is parameterized by Eq.~\ref{eq:oor_formulation}.

We first partition $\mathcal{P}$ into base object points $\mathcal{P}_{\mathcal{B}}$ and target object points $\mathcal{P}_{\mathcal{T}}$. Treating pseudo-multi-view images as a video and applying a video-segmentation model~\cite{ravi2024sam2segmentimages, ren2024grounding} yields better segmentation quality.
Next, we register each template mesh into the corresponding point cloud. For simplicity, we omit the object indicator ($\mathcal{B}$ or $\mathcal{T}$), and denote the point clouds as $\mathcal{P}$ and the template mesh in the canonical space as $\mathcal{M}$.
To register each mesh, we extract semantic correspondence features~\cite{zhang2024telling} from 2D views. Specifically, for each 3D point $\mathbf{P}_j$, its corresponding image projections $\{\mathbf{p}_j^k\}_{k=1}^{m_j}$ yield features $\mathbf{f}_j^k \in \mathbb{R}^F$ at each pixel, where $F$ is the semantic feature dimension ($F$=768 in practice). We apply the same process for each point on $\mathcal{M}$ from its rendered 2D views. 
Then we aggregate the semantic features from 2D views into a single feature vector for the corresponding 3D point. 
Rather than directly averaging these 2D features, we first run PCA~\cite{hotelling1933analysis} on all (pseudo-multi-view and rendered) features to obtain ${\mathbf{f}'}_j^k \in \mathbb{R}^{F'}$, where ${F'}<F$ ($F'$= 15 in practice). Finally, average them per 3D point.

Using the semantic features of $\mathcal{P}$ and $\mathcal{M}$, we establish correspondences by identifying nearest neighbors based on cosine similarity. We retain only those whose similarity exceeds a certain threshold.
We use Procrustes analysis~\cite{gower1975generalized, goodall1991procrustes} with RANSAC~\cite{fischler1981random} for the registration, followed by ICP~\cite{121791} to further refine it. To this end, we obtain the transformation $T$ from $\mathcal{M}$ to $\mathcal{P}$:
\begin{gather}\label{eqn:procrustes}
    T\left( \mathcal{M} ; s, \mathbf{R}, \mathbf{t} \right) = s \cdot \mathbf{R}\mathcal{M} + \mathbf{t} \approx  \mathcal{P},
\end{gather}
where $T$ is parameterized by 3D rotation $\mathbf{R} \in SO(3)$, 3D translation $\mathbf{t} \in \mathbb{R}^3$, and an isotropic scaling factor $s \in \mathbb{R}_{+}$, transforming the point in the $\mathcal{M}$ into the point cloud space. 

Applying the registration to each template mesh gives $T_{\mathcal{B} \rightarrow \mathcal{P}}$ and $T_{\mathcal{T} \rightarrow \mathcal{P}}$, the transforms from the base and target meshes to the point cloud space. 
Finally, we can find the relative pose and scale of the $\mathcal{M}_{\mathcal{T}}$ relative to $\mathcal{M}_{\mathcal{B}}$, denoted as
$T_{\mathcal{T} \rightarrow \mathcal{B}}$:
\begin{equation}\label{eqn:relative_transform}
    \begin{gathered}
    T_{\mathcal{T} \rightarrow \mathcal{B}} = T_{ \mathcal{P} \rightarrow \mathcal{B}} \circ T_{\mathcal{T} \rightarrow \mathcal{P}},\\
        T_{\mathcal{T} \rightarrow \mathcal{B}} : \mathcal{M}_{\mathcal{T}} \mapsto s_{\mathcal{T} \rightarrow \mathcal{B}} \cdot \mathbf{R}_{\mathcal{T} \rightarrow \mathcal{B}} \mathcal{M}_{\mathcal{T}} + \mathbf{t}_{\mathcal{T} \rightarrow \mathcal{B}},
    \end{gathered}
\end{equation}
where $T_{ \mathcal{P} \rightarrow \mathcal{B}}$ is the inverse transformation of $T_{ \mathcal{B} \rightarrow \mathcal{P}}$.
Note that we use an isotropic scale $s \in \mathbb{R}_{+}$ during the registration by keeping the original aspect ratio. After the registration, we obtain the scale representation defined in Eq.~\ref{eq:oor_formulation}, as follows:
\begin{equation}\label{eqn:get_data}
    \begin{gathered}
      \mathbf{s}^{\mathcal{T} \rightarrow \mathcal{B}} := s_{\mathcal{T} \rightarrow \mathcal{B}} \cdot \text{BBOX}(\mathcal{T}), \\
      \mathbf{s}^{\mathcal{B}} := \text{BBOX}(\mathcal{B}).
    \end{gathered}
\end{equation}
Fig.~\ref{fig:db_gen} overviews the OOR dataset generation process. To account for the shape deviations, we use several template meshes as candidates and select the best via DINO features~\cite{oquab2023dinov2, darcet2023vitneedreg}. Further, we apply a series of automatic filtering processes to remove unreliable samples. See Sec.~\ref{sec:supp_de_gen} for details.

\subsection{OOR Diffusion}
\label{sec:diffusion}
\begin{figure}[t]
    \centering
    \includegraphics[width=\linewidth, trim={0 0 0 0},clip]{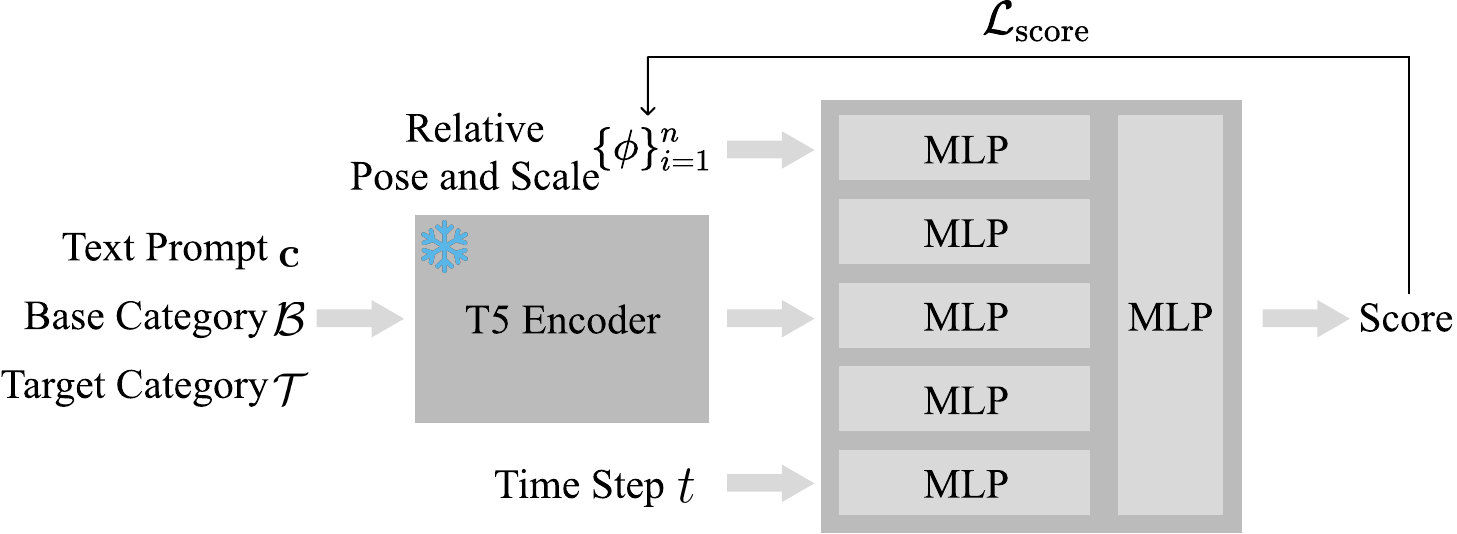}
    \captionof{figure}{\textbf{
    Training Process of OOR Diffusion.} Our OOR diffusion generates OOR samples by taking the context $\mathbf{c}$, base object category $\mathcal{B}$, and target object category $\mathcal{T}$ as text conditions.
    }
    \label{fig:diffusion}
    \vspace{-15pt}
\end{figure}

We model the pairwise OOR distribution $p_\mathbf{c}^{\mathcal{T} \rightarrow \mathcal{B}}$ via a diffusion model conditioned by various OOR contexts $\mathbf{c}$, with our collected dataset in Sec.~\ref{sec:dbgen}.
Our OOR diffusion model is based on a score-based model, following the approach of GenPose~\cite{zhang2024generative} that is originally introduced for 6D object pose estimation. Let our OOR diffusion $\Psi_\theta$, parameterized with $\theta$.
Then $\Psi_\theta$ models the noised score function at time step $t$ of the $p_\mathbf{c}^{\mathcal{T}\rightarrow \mathcal{B}}$:
\begin{equation}\label{eqn:scorenet}
    \begin{gathered}
       \Psi_\theta(\phi_{t}, t | \mathbf{c}, \mathcal{B}, \mathcal{T}) = \nabla_{\phi_{t}} \log p_\mathbf{c}^{\mathcal{T} \rightarrow \mathcal{B}}(\phi_{t}),
    \end{gathered}
\end{equation}
where $\phi_{t}$ is a noised OOR sample at time step $t$, $\mathbf{c}$ is the text prompt of OOR context, $\mathcal{B}$, and $\mathcal{T}$ represent base object category and target object category, respectively. Specifically, we take $\mathbf{c}$, $\mathcal{B}$, and $\mathcal{T}$ as text input and encode them with the pre-trained T5 text encoder~\cite{2020t5}.
According to Denoising Score Matching(DSM)~\cite{vincent2011connection}, by optimizing the following objective function, we can obtain 
Eq.~\ref{eqn:scorenet}:
\begin{equation}\label{eqn:score_loss}
    {\footnotesize
    \begin{gathered}
       \mathcal{L}_{\text{score}}(\theta) = \mathbb{E}_{t}\left[
            \lambda_{t}\mathbb{E}_{
                \phi, \phi_{t}
            }\left[ \left\| \Psi_\theta(\phi_{t}, t | \mathbf{c}, \mathcal{B}, \mathcal{T}) - \frac{\phi - \phi_{t}}{\sigma(t)^2} \right\|_2^2 \right]
        \right],
    \end{gathered}
    }
\end{equation}
where $t \sim (\epsilon, 1)$ with minimal noise level $\epsilon$, $\phi \sim p_\mathbf{c}^{\mathcal{T} \rightarrow \mathcal{B}}$, $\phi_{t} \sim \mathcal{N}(\phi, \sigma^2(t) \mathbf{I})$, $\sigma^2(t)$ is variance factor that increases exponentially with noise level, and $\lambda_{t}$ is regularization term according to noise level. The model architecture and training process of our OOR diffusion are shown in Fig.~\ref{fig:diffusion}.
For the inference, we can sample an OOR $\phi \sim p_\mathbf{c}^{\mathcal{T} \rightarrow \mathcal{B}}$, from pure Gaussian noise using the reverse ODE process~\cite{song2021scorebased, dormand1980family}, which solves the following Probability Flow ODE from $t=1$ to $t=\epsilon$:
\begin{equation}\label{eqn:reverse_ode}
    \begin{gathered}
       \frac{d\phi_{t}}{dt} = -\sigma(t)\dot\sigma(t)\nabla_{\phi_{t}} \log p_\mathbf{c}^{\mathcal{T} \rightarrow \mathcal{B}}(\phi_{t}).
    \end{gathered}
\end{equation}

\noindent \textbf{Text Context Augmentation.}
For better generalization, we leverage an LLM~\cite{openai2024chatgpt} to diversify the OOR context $\mathbf{c}$ in two ways. First, we augment $\mathbf{c}$ with variations in prompt phrasing (e.g., verbs and sentence structure), while preserving the original semantic meaning.
Second, we also diversify the object categories where the same relations hold and have the similar shape. For example, the context ``A teapot pours tea into a teacup" can represent similar spatial relations with ``A kettle pours tea into a cup''. We ask LLM to suggest alternative object categories with similar shapes and sizes, while preserving the same OOR distributions. Our text context augmentation improves generalization by enabling OOR modeling under a broader range of contextual conditions.
For details on data augmentation, see Sec.~\ref{sec:supp_diffusion}. As a result, we train our model for 475 distinct contexts with 23750 text prompts, each capturing various spatial relationships between object pairs across 188 object categories.

\begin{figure}[t]
    \centering
    \includegraphics[width=\linewidth, trim={0 0 0 0},clip]{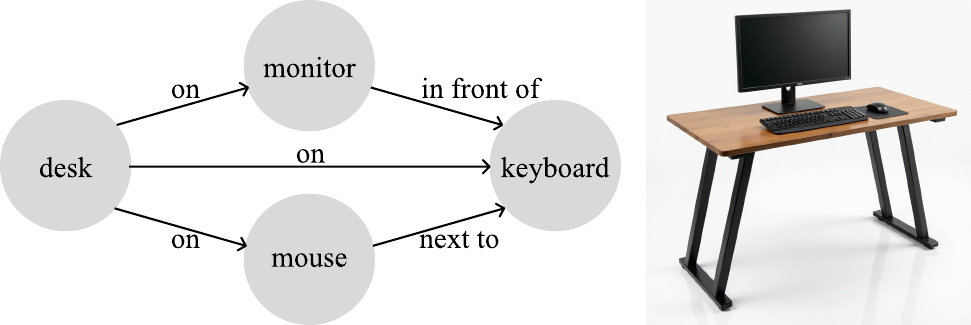}
    \captionof{figure}{\textbf{Scene Graph Example.} The scene graph for multi-object OOR is represented as a connected DAG with one start node.}
    \label{fig:scene_graph}
    \vspace{-15pt}
\end{figure}

\noindent \textbf{Extend to Multi-object OOR.}
To model the spatial relationships of multiple objects within a shared 3D space, we present a method to extend our pairwise OOR contexts into multi-object OORs. Given a set of pairwise contextual conditions describing a scene with multiple objects (e.g., Fig.~\ref{fig:scene_graph}), we aim to generate multi-object OORs from $n$ pairwise OOR distributions, $\{ \phi_{t}^{p_i} \}^n_{i=1} \sim \left\{p_i\right\}^n_{i=1}$, where $p_i$ represent the $i$-th pairwise OOR.
Specifically, we represent a scene as a connected Directed Acyclic Graph (DAG) with a single starting node (global base), as shown in Fig.~\ref{fig:scene_graph}. Each node represents an object, while each edge represents a pairwise OOR, where base-target object relationships are structured as parent and child nodes.
The final scene layout is determined by setting the global base node as the reference for the global coordinate system, and sequentially applying the sampled pairwise spatial relations($\phi_{t}^{p_i}$) following the graph path to position all objects within the global coordinate system. 

However, naively sampling from each pairwise OOR introduces two critical challenges: (1) collisions may occur between objects, (2) inconsistency issues arise when multiple pairwise relations define an object's pose and scale differently. For instance, in Fig.~\ref{fig:scene_graph}, the keyboard's spatial pose and scale can be determined via two distinct paths, relative to either the monitor or the mouse. Since these paths may yield different OOR layouts, a simple averaging of multiple available OOR cues may not produce a plausible final arrangement.

To address these issues, we present an approach to infer all multi-object OORs $\Phi = \left\{\phi_{t}^{p_i}\right\}_{i=1}^n$ simultaneously, by including our novel inference-loss terms: collision loss $C(\Phi)$ and inconsistency loss $I(\Phi)$.
This modifies the reverse ODE process in Eq.~\ref{eqn:reverse_ode} for all $i$ as follows: 
\begin{equation}\label{eqn:multi-oor}
    \begin{gathered}
       \frac{d\phi_{t}^{p_i}}{dt} = -\sigma(t)\dot\sigma(t)\nabla_{\phi_{t}^{p_i}} \log p_i(\phi_{t}^{p_i}) \\ + \lambda_1\nabla_{\phi_{t}^{p_i}}C(\Phi)
       +  \lambda_2\nabla_ {\phi_{t}^{p_i}}I(\Phi),
    \end{gathered}
\end{equation}
where $\lambda_1$ and $\lambda_2$ are weight terms.
The collision loss penalizes overlapping axis-aligned bounding boxes of positioned objects based on the sampled OOR cues. Note that the collision loss is only applied to object pairs without an explicitly defined OOR(i.e., two non-adjacent nodes) in the scene. In other words, for each object pair without an explicitly defined OOR, we assume that their spatial relationship corresponds to a non-colliding configuration.
The inconsistency loss minimizes the variance among OOR cues for the same object from different base object paths. See Sec.~\ref{sec:supp_diffusion} for details of inconsistency loss.

%% file: sec/4_experiments.tex
\begin{figure*}[t]
    \centering
    \includegraphics[width=\linewidth, trim={0 0 0 0},clip]{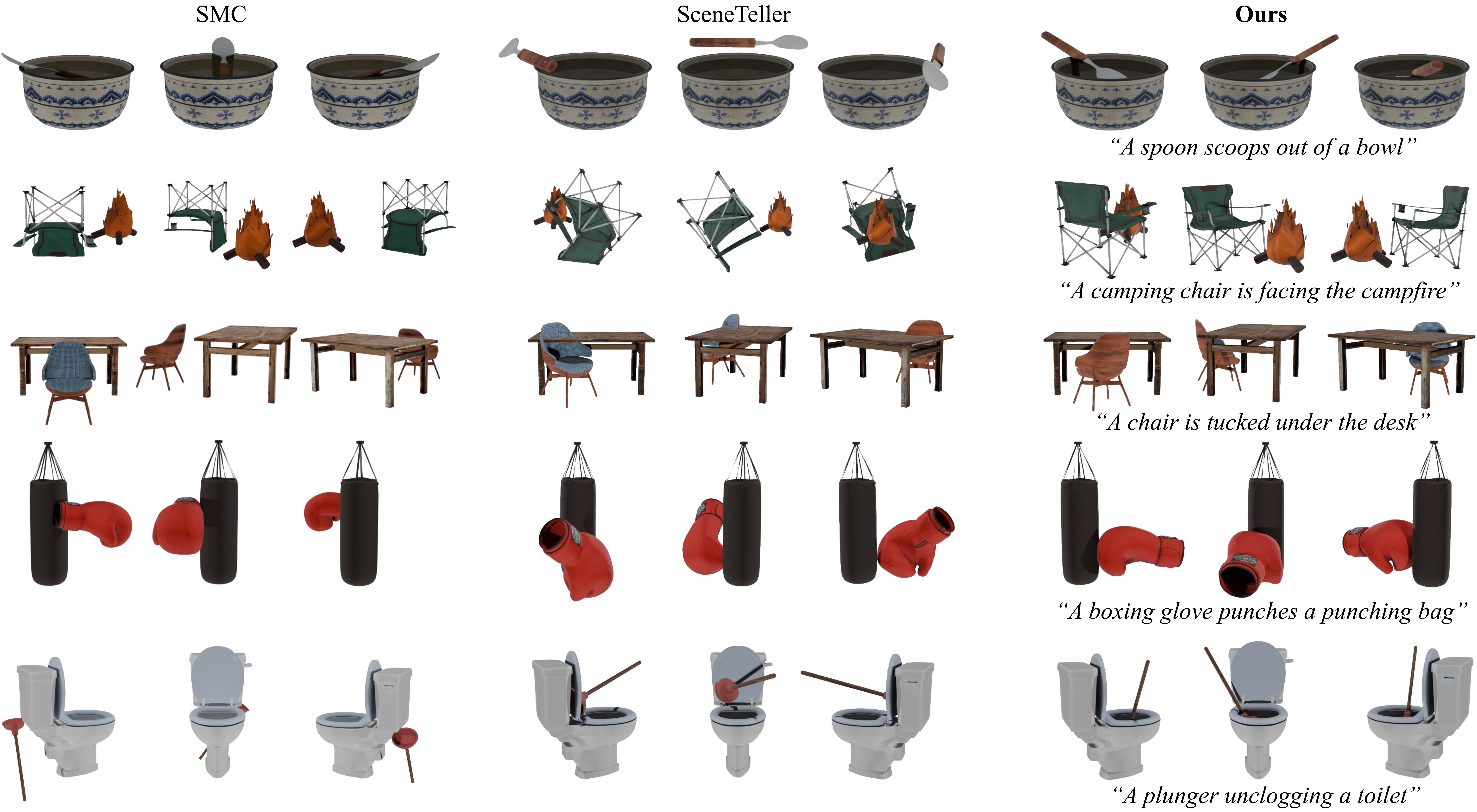}
    \captionof{figure}{\textbf{
    Qualitative Comparisons of Pairwise OOR Generation.}
    Our method models object-object spatial relationships better than LLM-based baselines.}
    \label{fig:qual_single}
    \vspace{-10pt}
\end{figure*}

\section{Experiments}
\label{sec:experiments}

In this section, we demonstrate the practical efficacy and generalization of our method. In Sec.~\ref{sec:single_oor}, we show that our diffusion model generates OORs that fit the text context more plausibly and effectively than other methods. Sec.~\ref{sec:multi_oor} demonstrates our advanced sampling approach produces significantly better results compared to text-to-3D models. In Sec.~\ref{sec:applications}, we present several applications that utilize our OOR. These are based on an optimization technique that leverages the characteristics of the score-based model.

\subsection{Pairwise OOR Generation}
\label{sec:single_oor}

\noindent \textbf{Baseline Methods.}
Since the concept of OOR is new in the field, there is no existing competitor with the same goal. Thus, we compare ours with LLM-based methods that pursue similar goals: SceneMotifCoder (SMC)~\cite{tam2024smc}, which focuses on 3D object alignments, and SceneTeller~\cite{ocal2024sceneteller}, which deals with 3D layout on a plane.

\input{tabs/single_oor.tex}

\noindent \textbf{Metrics.} 
Since there is no existing metric for OOR, we introduce several evaluation metrics to assess text prompt alignment with the plausibility of spatial relationship outputs in multi-view renderings. The CLIP score~\cite{radford2021learning} measures text-image alignment by averaging CLIP model logits. The VQA score~\cite{lin2025evaluating} leverages a VQA model to assess object composition and relations. 
However, since these methods evaluate per single image, we adopt GPTEval3D~\cite{wu2023gpteval3d} to evaluate the text-to-3D alignment in multi-view images by leveraging VLMs~\cite{openai2024chatgpt}. Inspired by this, we propose the VLM score, which evaluates how well OOR is represented in multi-view images. The VLM score is measured in a pairwise manner, averaging the percentage of times a method is preferred (with a maximum score of 100), while allowing ties. Additionally, we conduct a user study on CloudResearch, surveying a total of 92 respondents.

\noindent \textbf{Results.} 
We evaluate our method and the baselines on a total of 150 scenes derived from 30 category pairs with 5 scenes generated per prompt. Each scene is rendered into $10$ views with a white background. SMC outputs the arranged meshes, while SceneTeller provides the relative pose and scale. Both baselines handle simple relations such as ``on" or ``next to" reasonably, but they struggle with functional relationships. As shown in Fig.~\ref{fig:qual_single}, SMC often completely misinterprets rotation information despite handling translation appropriately. On the other hand, SceneTeller benefits from the strong in-context learning capability of LLMs, enabling reasonable estimations. However, due to the inherent limitation of estimating 3D information without direct 3D data, it lacks fine-grained control. In contrast, our OOR diffusion demonstrates superior sampling capabilities compared to the baselines, leveraging its effective learning of the OOR distribution with respect to the text context. Tab.~\ref{tab:single-oor} shows that our method outperforms baselines for all metrics.

\begin{figure}[t]
    \centering
    \includegraphics[width=\linewidth, trim={0 0 0 0},clip]{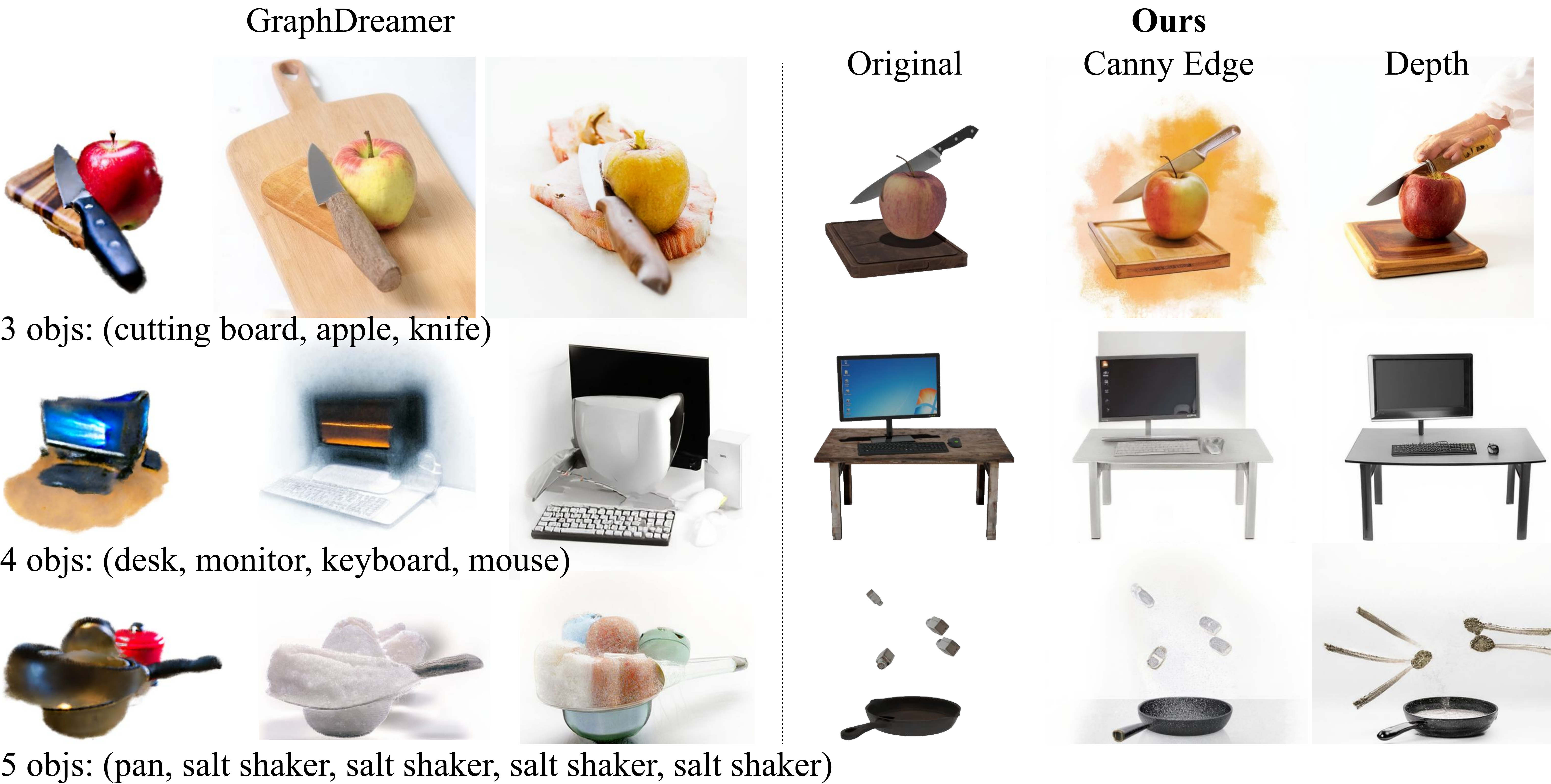}
    \captionof{figure}{\textbf{
    Qualitative Comparisons of Multi-object OOR Generation.} We generate multi-object OOR by combining each sample from our pairwise OOR diffusion model. Our method better captures multi-object OOR compared to the diffusion-based text-to-3D model.
    }
    \label{fig:multi-oor}
    \vspace{-10pt}
\end{figure}

\subsection{Multi-object OOR Generation}
\label{sec:multi_oor}

\noindent \textbf{Implementation Details.}
The specific settings for the weight terms in Eq.~\ref{eqn:multi-oor} are as follows:
(1) $\lambda_1 = \min(\frac{100}{t}), 10^4)$; (2) $\lambda_2 = \min(\frac{100}{t^2}, 10^5)$.
Additionally, this constraint is applied starting from $t = 0.5$.

\noindent \textbf{Baseline Method.} 
Since SMC and SceneTeller cannot be directly extended to multi-object OOR using only pairwise OOR data, we compare our model to another baseline GraphDreamer~\cite{gao2024graphdreamer}, which is diffusion-based text-to-3D model. While GraphDreamer employs neural implicit representation~\cite{wang2021neus} for 3D objects, which may result in low texture quality. Since our method renders scenes with template meshes, direct comparison is challenging. To mitigate the impact of rendering quality, we use ControlNet~\cite{zhang2023adding} to standardize the output style of synthesized images by using Canny edge or depth as guidance.

\noindent \textbf{Metrics.}
We use the same metrics as in Sec.~\ref{sec:single_oor}. However, for per-image scores such as CLIP score and VQA score, we apply them only to the ControlNet-generated images. Meanwhile, for multi-view-dependent metrics such as VLM score and user study(conducted with 81 respondents), we evaluate the original renderings. We guide both the VLM and human evaluators to ignore rendering quality and focus primarily on OOR. 

\input{tabs/multi_oor.tex}

\noindent \textbf{Results.}
We evaluate 20 scenes where 3 to 5 objects have spatial relations with each other. As shown in Fig.~\ref{fig:multi-oor}, GraphDreamer often fails to capture OOR (e.g., ``A knife cuts an apple."). It even omits certain objects, such as a computer mouse or a salt shaker. In contrast, our method reliably generates multi-object OOR by leveraging pairwise OOR knowledge. The quantitative comparison in Tab.~\ref{tab:multi-oor} further demonstrates the superiority of our method, especially in the case of VLM score and user study.

\begin{figure*}[t]
    \centering
    \includegraphics[width=\linewidth, trim={0 0 0 0},clip]{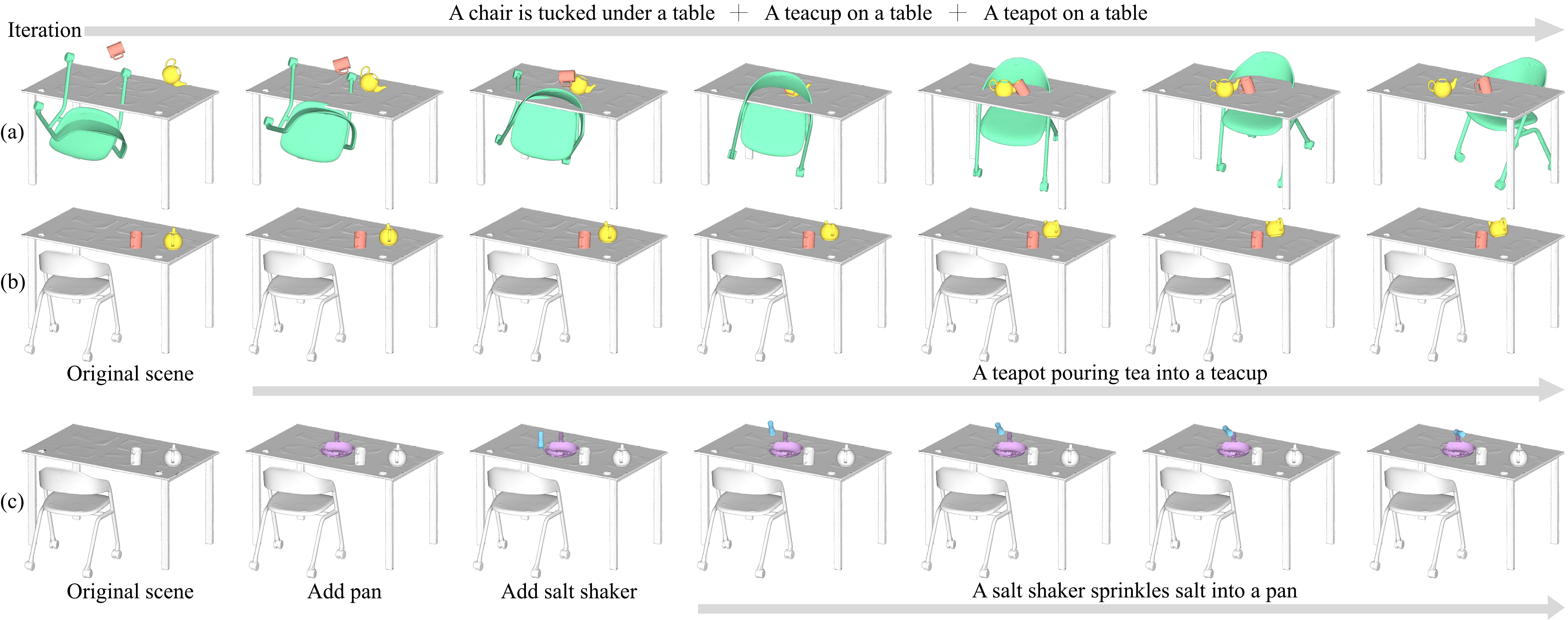}
    \captionof{figure}{\textbf{
    Scene Editing Results.}
    Results of editing the original ParaHome scene, which includes a table, chair, teacup, and teapot, using our OOR diffusion. \textbf{(a)} adding random noise to the original scene and then rearranging it. \textbf{(b)} applying the context ``A teapot pouring tea into a teacup" to the original scene. \textbf{(c)} adding a pan and a salt shaker to the original scene and applying ``A salt shaker sprinkles salt into a pan."}
    \label{fig:scene_edit}
    \vspace{-10pt}
\end{figure*}

\subsection{Applications of OOR}
\label{sec:applications}

\noindent \textbf{3D Scene Editing.}
Our OOR diffusion, built on a score-based model, enables the application of optimization techniques in various 3D scene editing scenarios.
First, we present a method that denoises a scene with multiple objects in a noisy arrangement, producing a more plausible and realistic configuration by guiding it toward high-likelihood regions in the OOR distribution.
For simplicity, consider a scene with a pair of objects, a base object and a target object, $(\mathcal{B}, \mathcal{T})$, with a relationship defined by context $\mathbf{c}$.
Given a noisy scene arrangement $\phi_0$ as input, our method optimizes the OOR $\phi \sim p_\mathbf{c}^{\mathcal{T} \rightarrow \mathcal{B}}$ by minimizing the following objective function:
\begin{equation}\label{eqn:objective}    
    \begin{gathered}
       \mathcal{L}_{\text{arrange}}(\phi) = ||\phi - \phi_0||^2_2 - \lambda_1 \log p_\mathbf{c}^{\mathcal{T} \rightarrow \mathcal{B}}(\phi)
    \end{gathered},
\end{equation}
where $\lambda_1$ is a weight term.
By minimizing Eq.~\ref{eqn:objective}, our method projects the input arrangement $\phi_0$ into a higher-likelihood region under $p_\mathbf{c}^{\mathcal{T} \rightarrow \mathcal{B}}$. The gradient of the objective function is:
\begin{equation}\label{eqn:gradient}    
    \begin{gathered}
       \nabla_\phi \mathcal{L}_{\text{arrange}}(\phi) = 2(\phi - \phi_0) - \lambda_1 \nabla_\phi \log p_\mathbf{c}^{\mathcal{T} \rightarrow \mathcal{B}}(\phi)
    \end{gathered}.
\end{equation}
Using Eq.~\ref{eqn:scorenet}, we can approximate the gradient of log-likelihood term as, $ \Psi_\theta(\phi, \epsilon | \mathbf{c}, \mathcal{B}, \mathcal{T}) \approx \nabla_\phi \log p_\mathbf{c}^{\mathcal{T} \rightarrow \mathcal{B}}(\phi)$. Thus, we update $\phi$ with the following gradient descent:
\begin{equation}\label{eqn:gradient_descent}    
    \begin{gathered}
        \phi \leftarrow \phi - \eta[(\phi - \phi_0) - \lambda_1 \Psi_\theta(\phi, \epsilon | \mathbf{c}, \mathcal{B}, \mathcal{T})]
    \end{gathered},
\end{equation}
where $\eta=0.01$ and $\lambda_1 = 0.01$.
Since the scale of each object is predetermined here, we only need to update the relative pose $\phi' = 
    (\mathbf{R}^{\mathcal{T} \rightarrow \mathcal{B}}, \mathbf{t}^{\mathcal{T} \rightarrow \mathcal{B}})$.
For multi-object scenarios, we leverage the introduced loss terms, $C(\Phi)$ and $I(\Phi)$.
In practice, optimization is completed within $50$ steps.
Also, we use our proposed optimization to apply different scene contexts $\mathbf{c}$ to a given arrangement. This can be performed by replacing the input context $\mathbf{c}$ in our OOR diffusion and optimizing with Eqs.~\ref{eqn:objective}--~\ref{eqn:gradient_descent}. Moreover, we add new objects into the scene by performing multi-object OOR generation (Eq.~\ref{eqn:multi-oor}) while keeping the OORs in the existing scene and the scale of the inserted objects fixed.

We demonstrate the efficacy of our proposed scene arrangement methods using ParaHome DB~\cite{kim2024parahome}, which provides 3D scenes with separate object meshes. Fig.~\ref{fig:scene_edit}(a) shows the result of rearranging (denoising) a scene after adding random noise to the original scene. In Fig.~\ref{fig:scene_edit}(b), we apply a scene context, where the teapot, which is initially placed next to the teacup, is adjusted to pour tea into the teacup. Finally, in Fig.~\ref{fig:scene_edit}(c), we demonstrate adding a pan and a salt shaker which are not originally present in the table, and then applying an optimization with a different scene context: ``A salt shaker sprinkles salt into a pan".

\begin{figure}[t]
    \centering
    \includegraphics[width=\linewidth, trim={0 0 0 0},clip]{figs/HOI_motion_gen.pdf}
    \captionof{figure}{\textbf{
    Human Motion Synthesis Results.}
    We assume two objects and a human interacting initially, and synthesize human motion using the generated OOR and contact consistency.}
    \label{fig:hms}
    \vspace{-10pt}
\end{figure}

\noindent \textbf{Human Motion Synthesis for Two-Object Interaction.}
As mentioned in Sec.~\ref{sec:intro}, OOR often reflects how humans interact with objects. We demonstrate this by synthesizing human motion interacting with two objects using generated OORs. First, we begin with the initial scene with two objects and the human interacting with them. For instance, a teapot and a teacup are placed on a desk, and the human has just started to grasp the teapot’s handle. By applying Eq.~\ref{eqn:gradient_descent}, we obtain the sequence of OOR $(\mathbf{R}^{\mathcal{T} \rightarrow \mathcal{B}}, \mathbf{t}^{\mathcal{T} \rightarrow \mathcal{B}})$ that takes it from the initial state to the OOR context $\mathbf{c}$. In the above example, setting $\mathbf{c} =$ ``A teapot pours tea into a teacup'', $\mathcal{B} =$ ``teacup'', and $\mathcal{T} =$ ``teapot'' in Eq.~\ref{eqn:gradient_descent} yields a sequence where the teapot, initially placed beside the teacup, moves to pour tea into it. Then, we can synthesize human motion by applying constraints to maintain initial contact between the human and the objects across entire OOR sequence. For each state in the OOR sequence, the human pose is obtained by optimizing the following objective function:
\begin{equation}\label{eqn:hms}
    {\footnotesize
    \begin{gathered}
       \mathcal{L}_{\text{hms}}(\theta_{\text{VP}}) = \lambda_1 ||\theta_{\text{VP}}||^2 + \lambda_2 \sum \mathbf{A}(\theta_{\text{VP}}) + \lambda_3 \mathbf{C}(\mathcal{H}, \mathcal{B}, \mathcal{T}),
    \end{gathered}
    }
\end{equation}
where $\theta_{\text{VP}}$ is a human body pose embedding of VPoser~\cite{SMPL-X:2019}, $\mathbf{A}$ is an angle prior of VPoser, $\mathcal{H}$ is a human derived from $\theta_{\text{VP}}$, $\mathbf{C}$ is a contact constraint between $(\mathcal{H}, \mathcal{B}, \mathcal{T})$, and $\lambda_i$ are weight terms. $\mathbf{C}$ maintains the distances of the $N$ nearest vertex pairs between the two object meshes and the human mesh at the initial state throughout the entire OOR sequence. Fig.~\ref{fig:hms} shows human motion interacting with two objects synthesized by this optimization.

%% file: tabs/single_oor.tex
\begin{table}[t]
    \centering
    \small{
        \begin{tabular}{l|ccc}
            \hline
            Metrics & SMC~\cite{tam2024smc} & SceneTeller~\cite{ocal2024sceneteller} & \textbf{Ours}  \\
            \hline
            CLIP Score $\uparrow$        & 28.54  & 29.06 & \textbf{29.11}  \\
            VQA Score $\uparrow$          & 0.61 & 0.68 & \textbf{0.69}  \\
            VLM Score $\uparrow$  & 49.83 &   64.67    &   \textbf{75.67}     \\
            User Study(\%) $\uparrow$  & 22.21   &  23.77   &   \textbf{54.02}   \\
            \hline
        \end{tabular}
    }
    \caption{\textbf{Quantitative Comparisons of Pairwise OOR Generation.} For each method, we evaluate CLIP score~\cite{radford2021learning}, VQA score~\cite{lin2025evaluating}, our proposed VLM score~\cite{wu2023gpteval3d}, and user study.}
    \label{tab:single-oor}
    \vspace{-15pt}
\end{table}

%% file: tabs/multi_oor.tex
\begin{table}[t]
    \centering
    \resizebox{0.5\textwidth}{!}{
        \begin{tabular}{l|cc|cc}
        \rowcolor{gray!20}
            \hline
            & \multicolumn{2}{c|}{Canny Edge ControlNet} & \multicolumn{2}{c}{Depth ControlNet} \\
            \hline
            Metrics & GraphDreamer~\cite{gao2024graphdreamer}  & \textbf{Ours} & GraphDreamer  & \textbf{Ours} \\
            \hline
            CLIP Score $\uparrow$        & 27.09  & \textbf{28.73} & \textbf{28.51} & 28.47 \\
            VQA Score $\uparrow$          & 0.44 & \textbf{0.71} & 0.62  & \textbf{0.79} \\
            \rowcolor{gray!20}
            \hline
            & \multicolumn{4}{c}{Original Rendering} \\
            \hline
            Metrics & \multicolumn{2}{c}{GraphDreamer} & \multicolumn{2}{c}{\textbf{Ours}} \\
            \hline
            VLM Score $\uparrow$          & \multicolumn{2}{c}{2.50} & \multicolumn{2}{c}{\textbf{97.50}}  \\
            User Study(\%) $\uparrow$          & \multicolumn{2}{c}{11.88} & \multicolumn{2}{c}{\textbf{88.12}}  \\
            \hline
        \end{tabular}
    }
    \caption{\textbf{Quantitative Comparisons of Multi-OOR Generation.}}
    \label{tab:multi-oor}
    \vspace{-20pt}
\end{table}

%% file: sec/5_conclusion.tex
\section{Discussion}
\label{sec:conclusion}
We present a novel approach to learning object-object spatial relationships (OOR) from synthetic 3D samples, leveraging pre-trained 2D diffusion models. We formulate a novel concept of OOR, and introduce a comprehensive pipeline for synthesizing 3D OOR samples, which can be applicable for any unbounded object categories. We develop a score-based diffusion model specifically designed to model 3D spatial relationships between objects. Furthermore, we demonstrate the scalability of our method by proposing a method to generate multi-object OOR by extending our pairwise OOR models. We also demonstrate the potential of our method for several applications by showing that scene editing and human motion synthesis can be effectively implemented via score-based updates. As a future direction, more factors, such as detailed object shapes, can be considered as additional factors to determine OOR.

%% file: sec/10_supp.tex
\clearpage
\appendix

\section{Details on Dataset Generation}
\label{sec:supp_de_gen}

\begin{figure}[t]
    \centering
    \includegraphics[width=\linewidth, trim={0 0 0 0},clip]{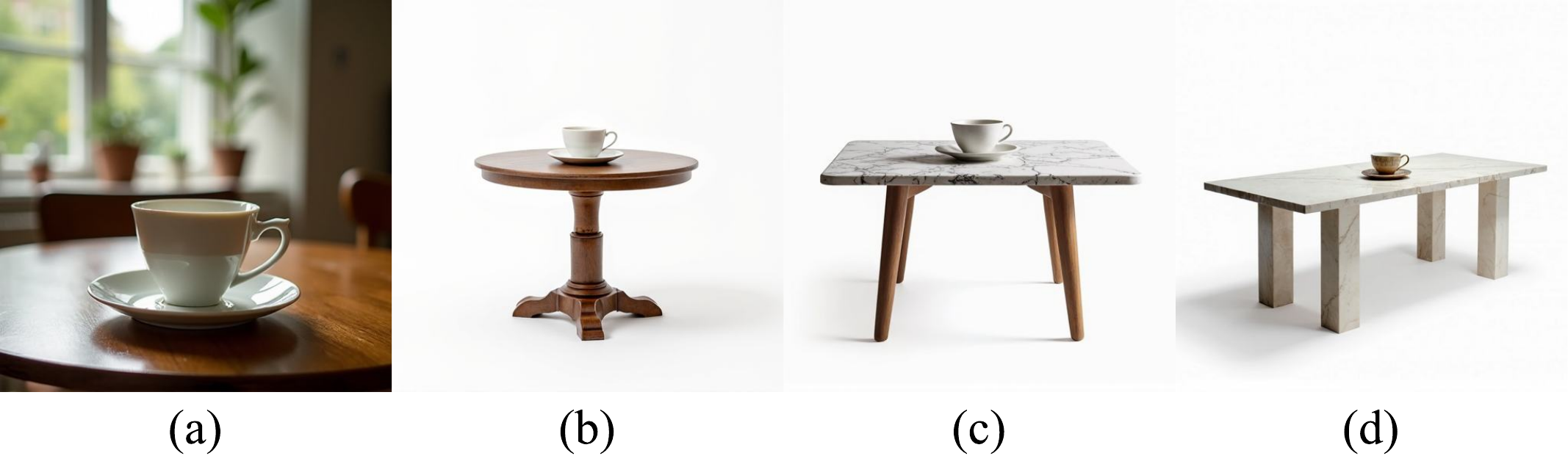}
    \captionof{figure}{\textbf{
    Controllability of Diffusion Models Through Text Prompts.} Each is an output image of diffusion generated with the following prompt: \textbf{(a)} ``A table with a teacup on top." \textbf{(b)} Adding ``White background." to the end of the prompt. \textbf{(c)} Specifying the shape and texture of the table as ``four-legged rectangular marble table." \textbf{(d)} Specifying the camera view via ``, from a diagonal angle."
    }
    \label{fig:prompt}
    \vspace{-10pt}
\end{figure}

\noindent \textbf{Text Prompts for Image Generation.}
The controllability of diffusion via text prompts offers advantages in learning OOR from synthetic images over real-world images by generating realistic OOR images while simultaneously enhancing their learnability through precise control. Fig.~\ref{fig:prompt} illustrates this clearly. (a) shows an image generated with a simple prompt: ``A table with a teacup on top.'' While the image is highly realistic, it poses challenges for learning the OOR between the ``table" and ``teacup" because the full shape of the table is not visible. (b) shows the result of adding ``White background.'' to the end of the prompt, which directs focus to the two objects and ensures that their full shapes are captured within the image frame without the need for additional context. (c) demonstrates control over the object's shape and texture in the generated image. This facilitates the registration of template object meshes. Finally, (d) shows how to mitigate frame size constraints by controlling the camera view. This is useful for capturing OOR between objects with large size differences, such as a table and a teacup, or between objects positioned at some distance, such as a monitor and a keyboard. We use the FLUX.1-dev~\cite{flux} in all image generation.

\begin{figure}[t]
    \centering
    \includegraphics[width=\linewidth, trim={0 0 0 0},clip]{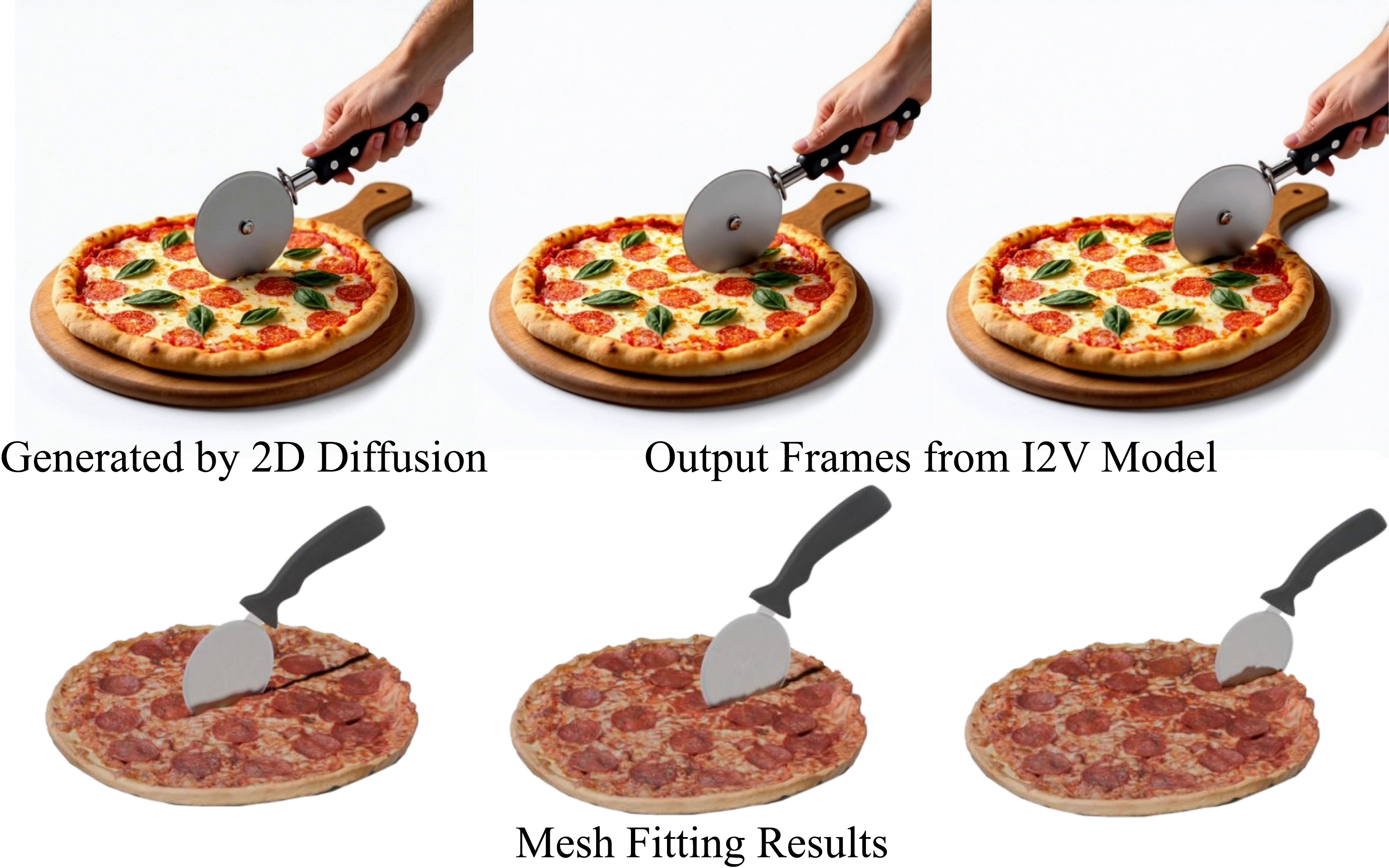}
    \captionof{figure}{\textbf{
    Image Augmentation via Image-to-Video Model.} We diversify scenes with dynamic object-object spatial relationships using the image-to-video model.
    }
    \label{fig:augmentation}
    \vspace{-10pt}
\end{figure}

\noindent \textbf{Synthetic Image Augmentation via Video Diffusion.}
We further augment 2D OOR images using the I2V model~\cite{KLING} for contexts where dynamic OOR can be generated by humans.
The motivation for using the image-to-video model is to generate a broader range of relative object relationships within each context, as image diffusion models typically produce the most representative configuration (e.g., the pizza cutter tends to be in the center of a pizza). We then use each frame from the synthesized videos as additional synthetic 2D samples, disregarding their temporal information. See Fig.~\ref{fig:augmentation} for the example result.

\noindent \textbf{Best Template Selection.}
If the shape of the object in the synthetic image and the template mesh are very different, mesh registration often fails. Therefore, for each category, we collect several template meshes as candidates and select the template that best matches the object in the image. To do this, we obtain DINO~\cite{oquab2023dinov2, darcet2023vitneedreg} features for $M$ pseudo multi-view images and $N$ mesh multi-view renderings, and select the mesh with the highest value by calculating the average of cosine similarity for $N \times M$ pairs. We collect template meshes from Sketchfab~\cite{sketchfab}. There are $96$ template meshes used for data generation.

\noindent \textbf{Filtering Process.}
Our filtering process is automated in the following steps: First, we filter out all the bad quality images where segmentation and SfM fail. In the process from SfM to feature matching, we filter out if the number of points corresponding to each base object and target object is less than $100$. The cosine similarity threshold is set to $0.7$ in most cases. There may still be misalignment between the registered mesh and the point clouds. We use the Chamfer Distance from the mesh to the point clouds. The threshold is adjusted according to the scale of the registered mesh. Most of the bad samples are filtered out through a series of processes, but some cases, such as flipped meshes, may remain. We use VLM~\cite{openai2024chatgpt} to filter out the last remaining bad samples. Specifically, we render combinations of base and target objects, and then ask VLM to judge whether the multi-view images align well with the text prompt, using the same criteria as when measuring the VLM score. The filtering ratio is $0.58$ to $0.92$.
However, since our approach is based on fully synthetic data, we can iterate this process as needed to obtain a sufficient number of high-quality 3D outputs. We obtain $30$ to $216$ samples per context.

\section{Details on OOR Diffusion.}
\label{sec:supp_diffusion}

Our OOR diffusion is trained for 20,000 epochs, taking about 10 hours on an RTX 6000 48GB.

\noindent \textbf{Architecture Implementation.}
We follow the implementation of ScoreNet in GenPose~\cite{zhang2024generative} for our score-based OOR diffusion. However, we take text as a condition instead of point clouds. For this, we introduce the T5 text encoder~\cite{2020t5}. Also, unlike Genpose, which only deals with the scores of rotation, translation, we also consider the 3-dimensional scale. For this, in the inference reverse ODE process, we add guidance to make the scale positive. Following GenPose, we consider a 6D representation~\cite{Zhou_2019_CVPR} for rotation. Therefore, our OOR diffusion is learned in a 15-dimensional space (6D target object rotation, 3D target object translation, 3D target object scale, 3D base object scale).

\begin{figure}[t]
    \centering
    \includegraphics[width=\linewidth, trim={0 0 0 0},clip]{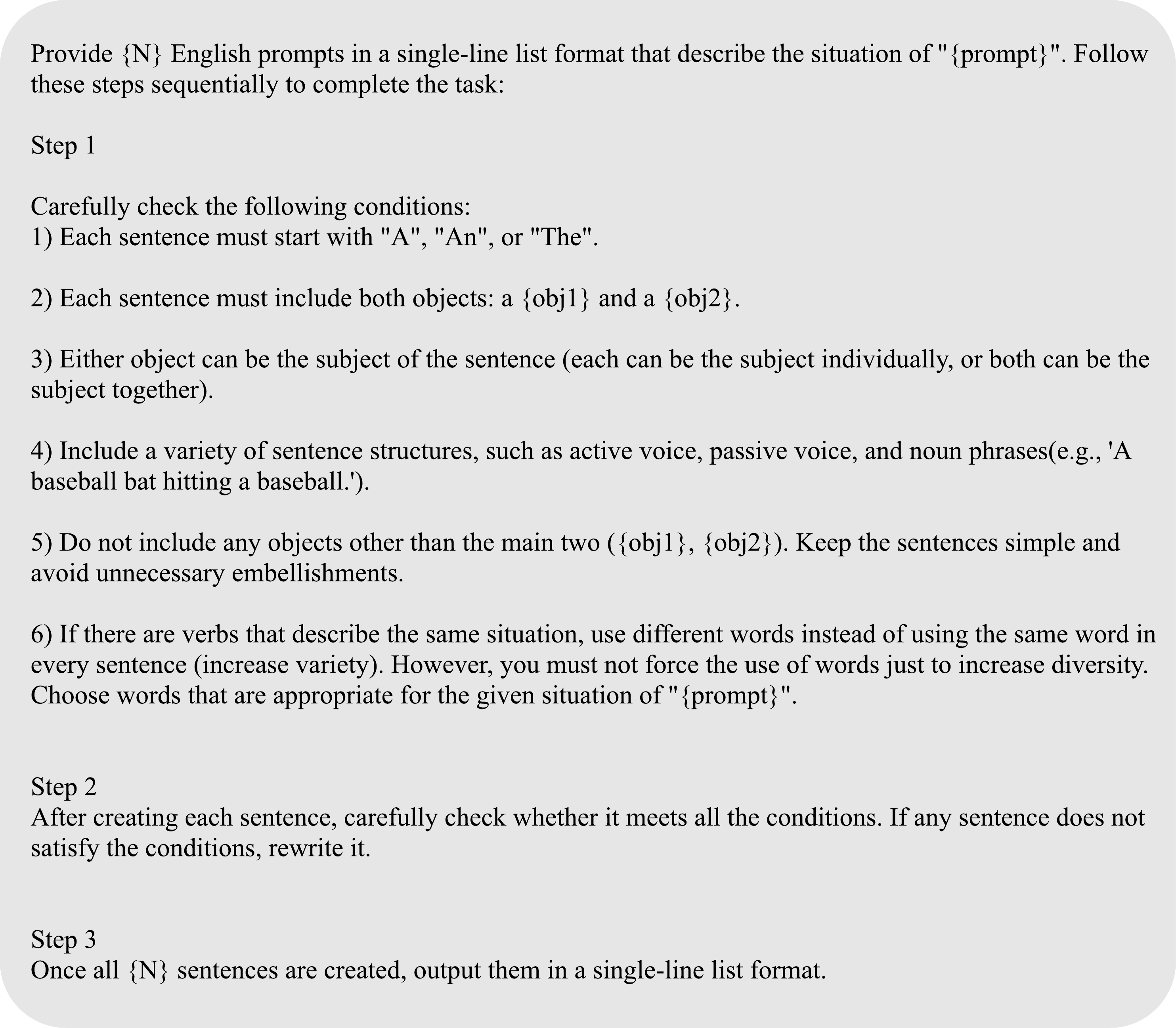}
    \captionof{figure}{\textbf{
    Guided Text prompt Provided to LLM for Text Context Augmentation.} LLM augments on text context $\mathbf{c}$ via the following guided prompt.
    }
    \label{fig:text_aug}
    \vspace{-10pt}
\end{figure}

\noindent \textbf{Text Context Augmentation.}
As proposed in Sec. \textcolor{iccvblue}{3}, we perform text context augmentation to increase the generality of OOR diffusion. Through the guided prompts in Fig.~\ref{fig:text_aug}, LLM generates various text prompts that describe a given context $\mathbf{c}$. Object categories are augmented by asking the LLM to present categories with similar shape and scale that could replace $\mathcal{B}$ and $\mathcal{T}$ in the given text context $\mathbf{c}$.

\noindent \textbf{Inconsistency Loss.}
The inconsistency loss introduced in Sec. \textcolor{iccvblue}{3.3} is computed as the average of the following three inconsistency parts:
(1) The scale variance of a global base object; (2) The pose and scale variance in the global coordinate system derived from different parents; (3) The variance of each component's ratio between the scale in the global coordinate system and the base scale in pairwise OOR, measured for parent nodes that are not global bases.
Specifically, (1) corresponds to the part related to the desk in Fig. \textcolor{iccvblue}{5}. OOR diffusion generates different $\mathbf{s}^{\mathcal{B}}$ for each pair, (desk, monitor), (desk, keyboard), and (desk, mouse), within the batch. In this case, (1) takes the variance of three $\mathbf{s}^{\mathcal{B}}$ as a loss term. (2) is the part corresponding to the keyboard in Fig. \textcolor{iccvblue}{5}. The pose and scale of each object in the global coordinate system are obtained as many times as the number of parent nodes of the corresponding node in the scene graph. Thus, the variance of three different poses and scales in global coordinate systems of the keyboard obtained from the desk, monitor, and mouse is the loss term in (2). (3) relates to the scale ratio consistency of monitor and mouse, which are parent nodes but not the global base. For example, the monitor should maintain consistency between its scale in the global coordinate system obtained from paths in the scene graph and the $\mathbf{s}^{\mathcal{B}}$ of the (monitor, keyboard) OOR sample. They do not have to be equal, but the ratio of each component should be constant. For example, if the obtained monitor scale in the global coordinate system is $(0.5, 0.4, 0.2)$, then the $\mathbf{s}^{\mathcal{B}}$ in the (monitor, keyboard) OOR sample should be $(1.0, 0.8, 0.4)$. The variance of the ratio of each component of the relevant scales is the loss term in (3).

\begin{figure}[t]
    \centering
    \includegraphics[width=\linewidth, trim={0 0 0 0},clip]{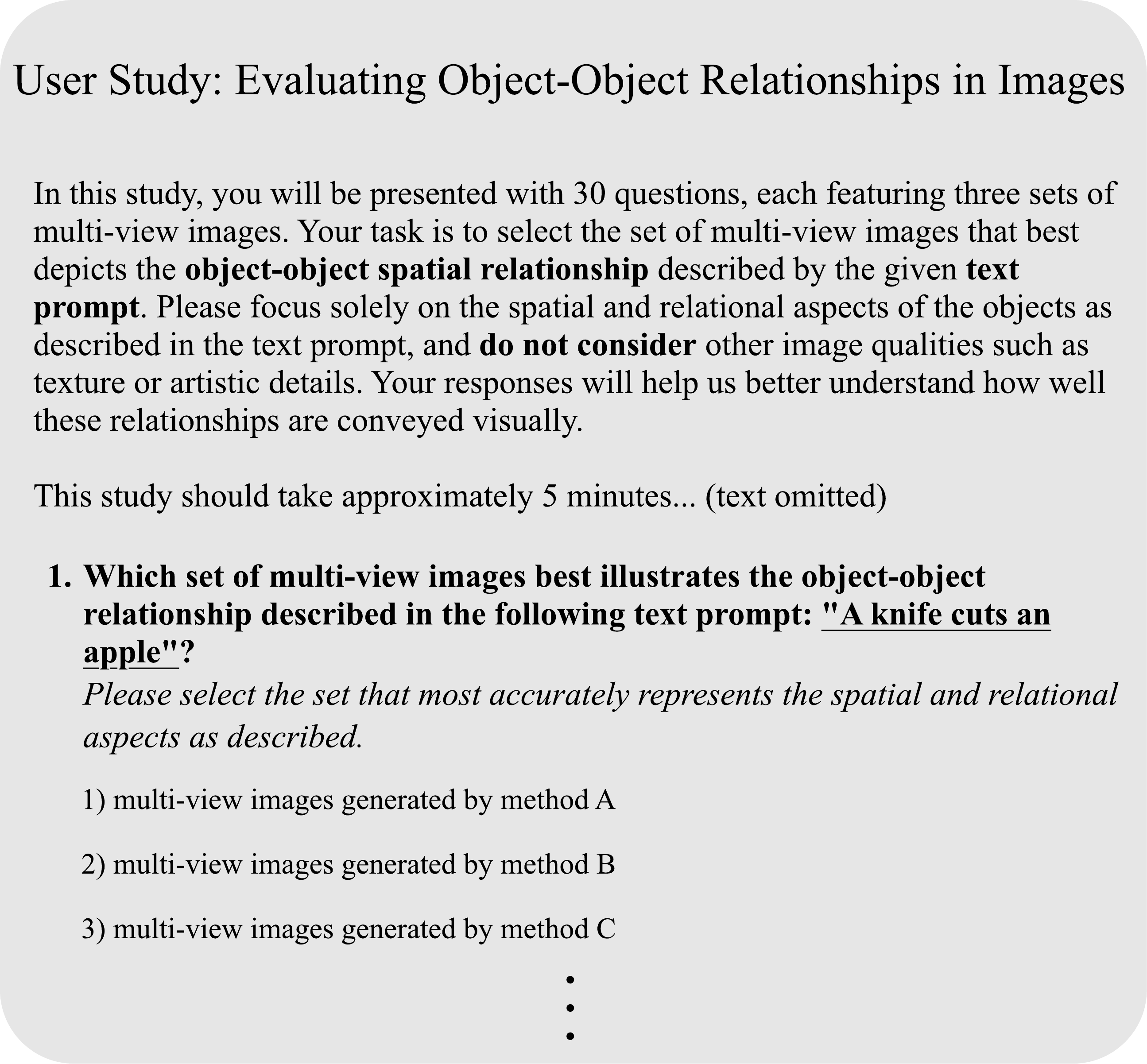}
    \captionof{figure}{\textbf{
    Questionnaire for User Study.} Participants select the multi-view image set that best captures the given OOR as instructed in the questionnaire.
    }
    \label{fig:user_study}
    \vspace{-10pt}
\end{figure}

\section{Experiments Details}
\label{sec:experiments_detail}

\noindent \textbf{Baseline Methods Details for Pairwise OOR Generation.}
SceneMotifCoder (SMC)~\cite{tam2024smc} is an example-driven visual program learning method. It takes text prompts as input and produces 3D object alignments by selecting and arranging meshes from a mesh pool. Given an example of a GT text prompt and mesh alignments in training, SMC analyzes patterns within the input, generates programs, and updates the program when new examples are introduced. During the inference process, when an input text comes in, LLM maps it to an appropriate task, and the program produces 3D object arrangements with the retrieved meshes from the candidate mesh pool. We convert our template mesh pairs obtained in the dataset generation process (Sec. \textcolor{iccvblue}{3.2}) into the SMC format. Since SMC is not concerned about the relative scale between objects, we use meshes with scales in our OOR dataset as a mesh pool during inference.

SceneTeller~\cite{ocal2024sceneteller} leverages LLMs for in-context learning by providing the LLM with pairs of (\textit{GT text prompt, GT scene layout}), enabling it to generate the appropriate scene layout for a test prompt. However, existing methods only focus on the layout for placement on the plane. To generate (\textit{GT text prompt, GT OOR}) pairs, we instruct the LLM with our world coordinate system and object canonical space. Then, we provide the GT OORs for generating corresponding text prompts. It also allows LLM to generate additional prompts for inference based on the generated GT prompts. For rotation, Euler angles format is chosen as the representation because LLMs tend to generate incomplete SO(3) matrices.

\begin{figure*}[t]
    \centering
    \includegraphics[width=17cm, trim={0 0 0 0},clip]{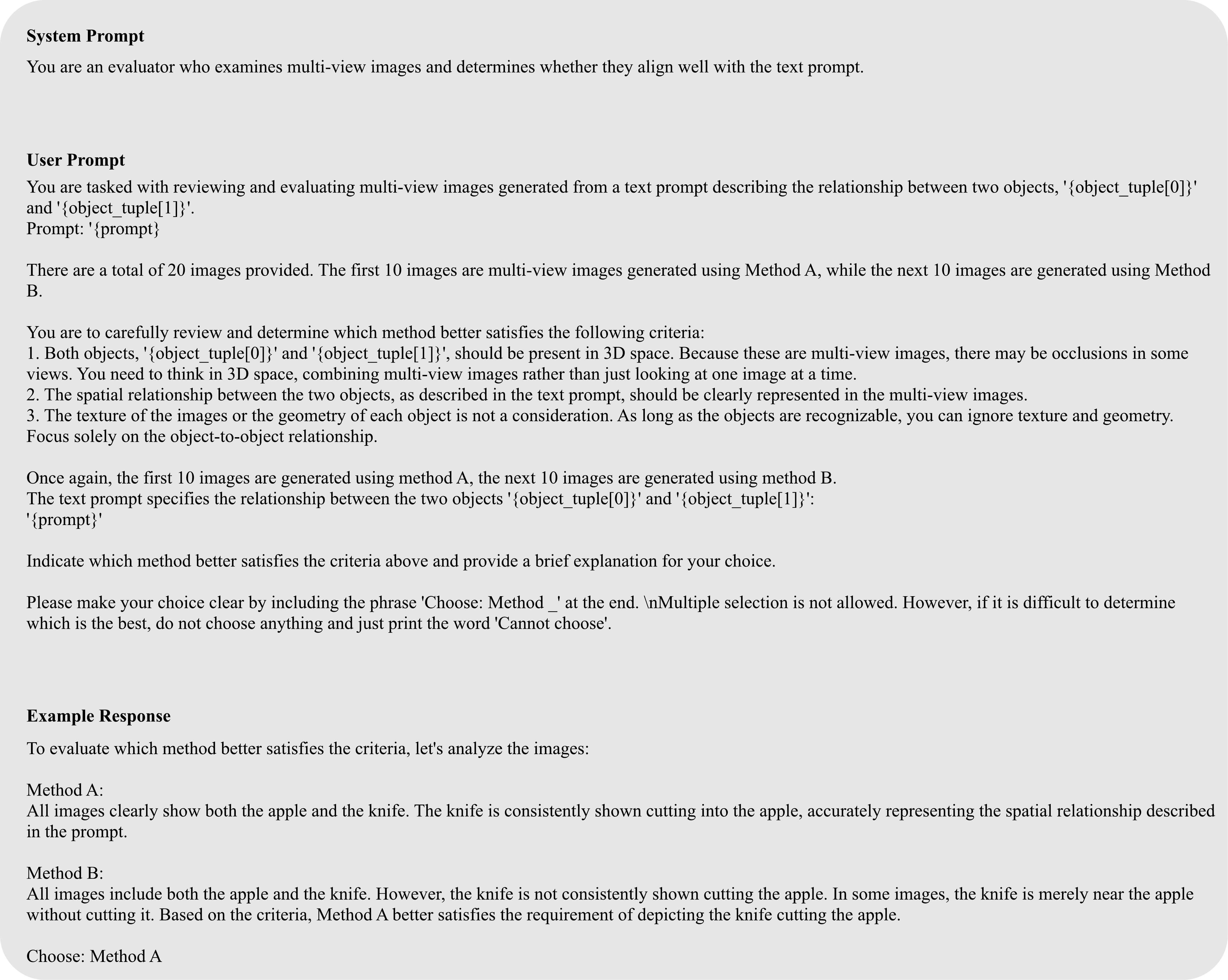}
    \captionof{figure}{\textbf{
    Guided Prompt Provided to VLM for VLM Score Evaluation.} Using the guided prompt above, VLM selects the preferred multi-view image set between the two generated by the different methods.
    }
    \label{fig:guided_prompt}
    \vspace{-10pt}
\end{figure*}

\noindent \textbf{VLM Score.}
As described in Sec. \textcolor{iccvblue}{4}, we propose the VLM score, inspired by GPTEval3D~\cite{wu2023gpteval3d}, to evaluate the alignment between the text context of OOR and multi-view images. We use VLM (specifically GPT-4o~\cite{openai2024chatgpt}) to compare two sets of multi-view images, each containing 10 images. These image sets are generated using our method or baselines. VLM is tasked with selecting the image set that better represents the spatial relationship between objects described in the text prompt. To ensure a fair comparison, we instruct VLM to ignore texture quality and focus solely on OOR. Fig.~\ref{fig:guided_prompt} illustrates the guided prompt we provided to VLM along with an example response from VLM.

\noindent \textbf{User Study.}
For the user study, we randomly select one scene from each of the object pairs per method to create a total of 30 questions for pairwise OOR generation. Similar to the VLM score evaluation, participants are instructed to disregard factors such as texture quality and focus solely on the OOR. For each question, multi-view image sets generated by methods A, B, and C are presented, and participants are asked to select the method that best represents the OOR described in the text prompt. To prevent bias, the order of A, B, and C is randomized for each question. We collect responses from 92 participants in total(81 participants for multi-OOR evaluation). The detailed questionnaire structure is illustrated in Fig.~\ref{fig:user_study}.

\begin{figure}[t]
    \centering
    \includegraphics[width=\linewidth, trim={0 0 0 0},clip]{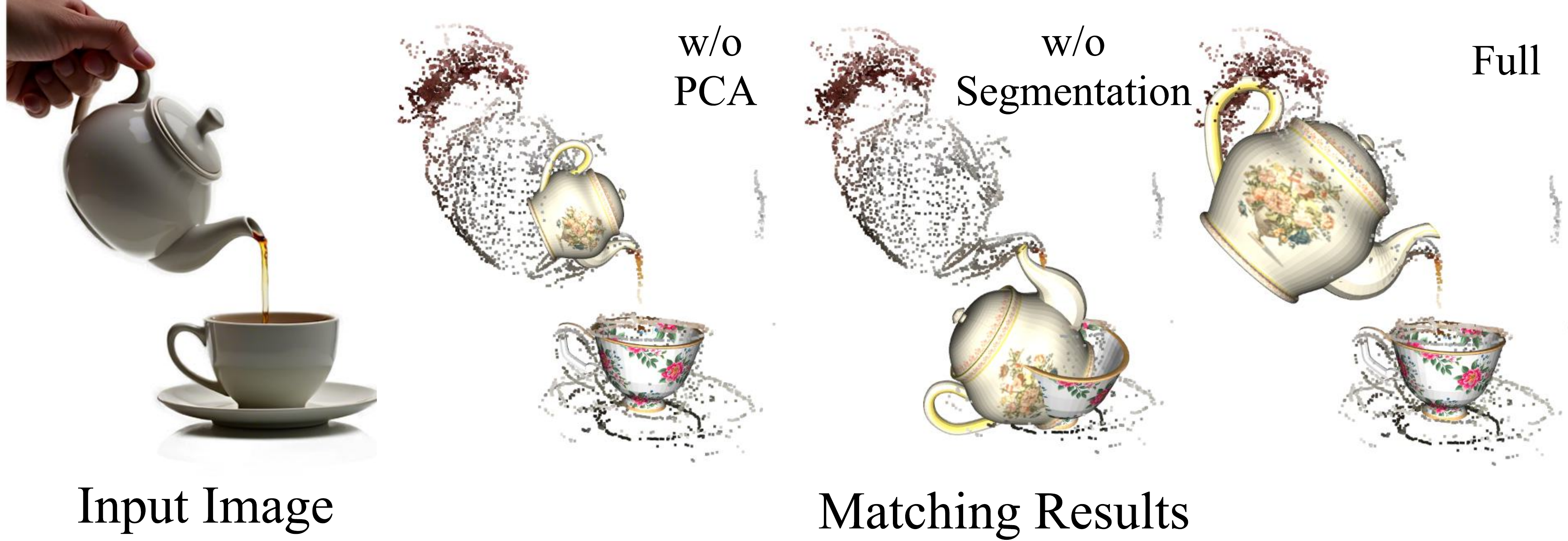}
    \captionof{figure}{\textbf{
    Ablation Study for Data Generation.}
    We show that applying PCA to features of points and separating the base object and target object through segmentation for matching in better results.}
    \label{fig:ablation}
    \vspace{-10pt}
\end{figure}

\input{supp_tabs/ablation.tex}

\section{Ablation Study}
\label{sec:ablation}
We compare our OOR distribution to the real data distribution and perform an ablation study to provide further justification for our OOR data generation pipeline.

\noindent \textbf{Dataset.}
We use the ParaHome DB~\cite{kim2024parahome}, which captures dynamic 3D movements of humans and objects in a home environment. We extract three OOR distributions: (`cutter board', `knife'), (`teacup, `teapot'), and (`pan', `salt shaker'). To exclude the approach and departure of humans relative to an object, we use the middle 70\% of sequences. Since there is only one instance for each category, the scale is constant. Therefore, we construct a joint distribution for $(\mathbf{R}^{\mathcal{T} \rightarrow \mathcal{B}}, \mathbf{t}^{\mathcal{T} \rightarrow \mathcal{B}})$.

\noindent \textbf{Baseline Methods.}
We ablate our mesh registration pipeline in Sec. \textcolor{iccvblue}{3.2} by removing point cloud separation and PCA on semantic features (SD+DINO~\cite{zhang2024telling}), comparing them with our full pipeline.

\noindent \textbf{Metric.}
We use the Fr{\'e}chet distance (FD)~\cite{dowson1982frechet} to measure distribution similarity.
When two distributions $p$ and $q$ are approximated by a multivariate Gaussian, the FD score $d$ of the two distributions is given by:
\begin{gather}\label{eqn:FD}
    d^2 = \|\mu_p - \mu_q\|^2 + \mathrm{tr}\left(\Sigma_p + \Sigma_q - 2\left(\Sigma_p \Sigma_q\right)^{1/2}\right),
\end{gather}
where $\mu_p$ is mean of $p$, $\mu_q$ is mean of $q$, $\Sigma_p$ is covariance matrix of $p$, and $\Sigma_q$ is covariance matrix of $q$.
Since rotation and translation have different units in each context, we train a 3-layer MLP encoder-decoder on 50M randomly sampled rotation matrix and translation vector. Then we compute FD in the learned 128D feature space.

\noindent \textbf{Results.}
Tab.~\ref{tab:ablation} shows that our method produces closer OOR distributions to real data than baselines, validating our full pipeline. Fig.~\ref{fig:ablation} further demonstrates the advantage of our segmentation and PCA modules. Without segmentation, registration often misaligns objects, and PCA enhances accuracy, yielding more realistic OOR samples.

\begin{figure}[t]
    \centering
    \includegraphics[width=\linewidth, trim={0 0 0 0},clip]{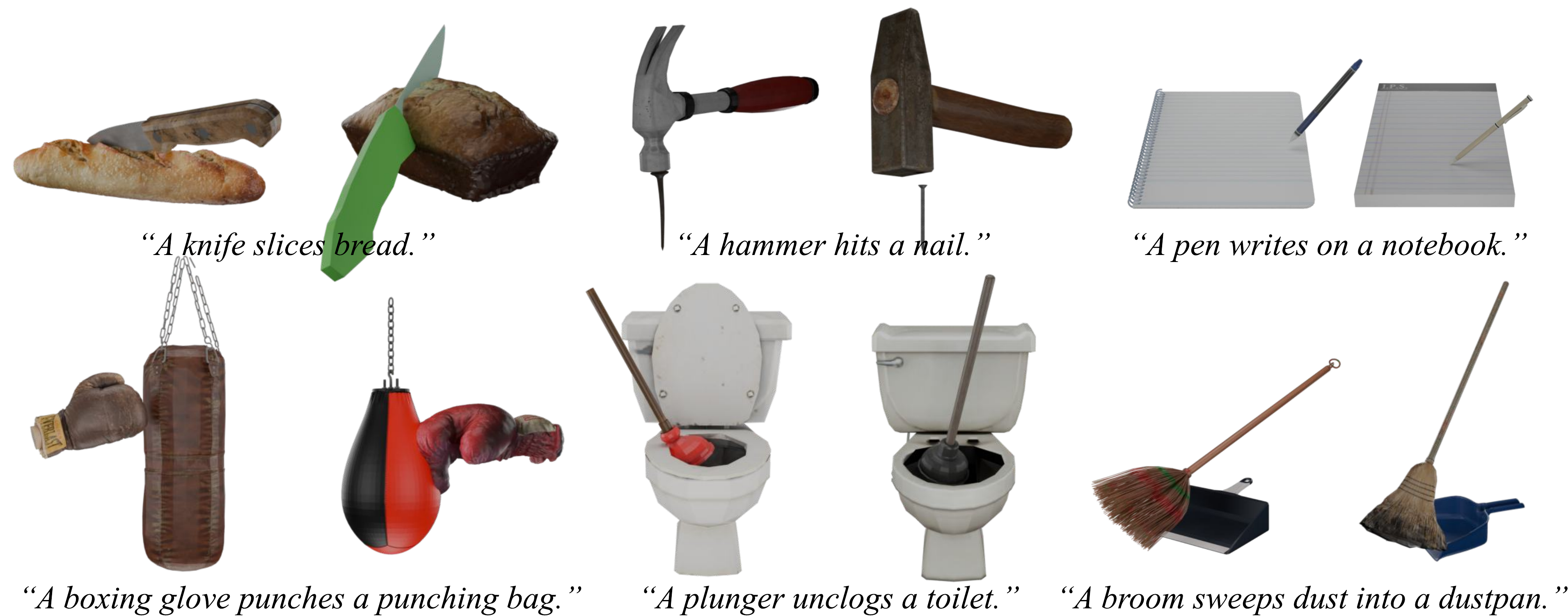}
    \captionof{figure}{\textbf{
    Applying Our OOR Diffusion Samples to Unseen Instances.} Our OOR diffusion still works when applied to instances other than the template meshes used to generate the dataset.
    }
    \label{fig:generality_mesh}
    \vspace{-10pt}
\end{figure}

\section{Generality of Our methods}
\label{sec:additional}

\noindent \textbf{Generality for Unseen Mesh Instances.}
Fig.~\ref{fig:generality_mesh} demonstrates the generality of our OOR modeling for unseen mesh instances. Our OOR diffusion generates appropriate relative poses and scales even for instances other than the template meshes used to generate the dataset. We consider the following scales to maintain the aspect ratio of each instance for both the base object and the target object:
\begin{gather}\label{eqn:isotropic}
    \mathbf{s}' := \text{Mean}(\mathbf{s} / \text{BBOX}(\mathcal{M})) \cdot \text{BBOX}(\mathcal{M}),
\end{gather}
where $\mathbf{s}$ is 3-dimensional scale from OOR diffusion, and $\mathcal{M}$ is an instance mesh.

\begin{figure}[t]
    \centering
    \includegraphics[width=\linewidth, trim={0 0 0 0},clip]{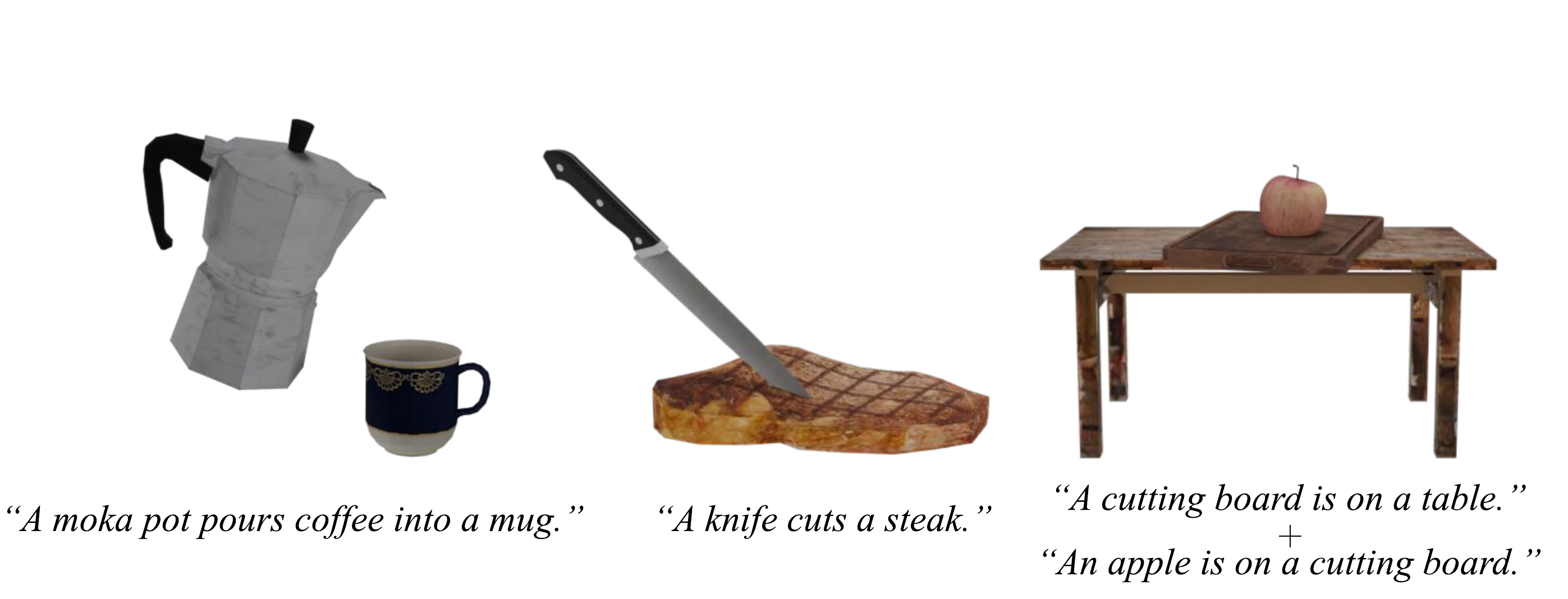}
    \captionof{figure}{\textbf{
    Our OOR Diffusion Sampling Results Under Unseen Text Prompt Condition.} Our OOR diffusion also works on text prompts that are not explicitly seen in training (including new categories and spatial relations).
    }
    \label{fig:generality_text}
    \vspace{-10pt}
\end{figure}

\noindent \textbf{Generality for Unseen Text Inputs.}
Fig.~\ref{fig:generality_text} shows that our OOR diffusion still produces plausible outputs even for text prompts that are not seen during training.
In the first example, the spatial relation ``pour" is learned, but the object categories ``moka pot" and ``mug" are not seen during training.
In the second example, both ``steak" and ``knife" are categories seen during training, but the spatial relationship of ``cutting steak with a knife" is not learned.
The last example shows a case of multi-object OOR. The spatial relation of placing a ``cutting board" somewhere is not seen during training, but thanks to the generality of OOR diffusion, it is correctly placed on the ``table".

\begin{figure}[t]
    \centering
    \includegraphics[width=\linewidth, trim={0 0 0 0},clip]{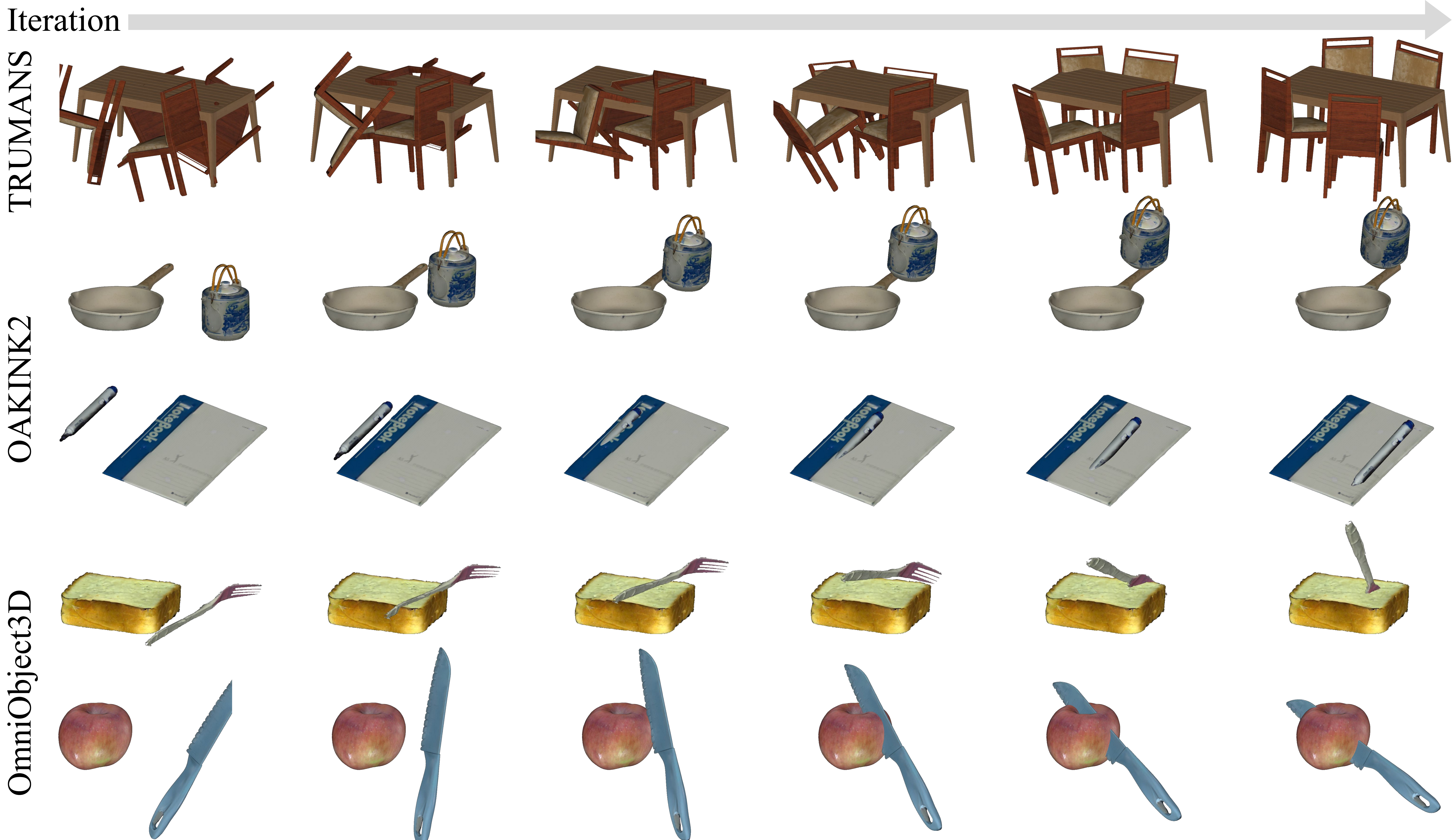}
    \captionof{figure}{\textbf{
    Scene Editing Results on TRUMANS, OAKINK2 and OmniObject3D.} Our scene editing algorithm works on a variety of real and synthetic datasets.
    }
    \label{supp_fig:scene_edit}
    \vspace{-10pt}
\end{figure}

\noindent \textbf{Generality for 3D Scene Editing.}
Since our OOR can be applied regardless of objects' textures or shape details, the synthetic-to-real gap is minimal.
Fig.~\ref{supp_fig:scene_edit} demonstrates the effectiveness of our scene editing algorithm on additional datasets which are TRUMANS, OAKINK2, and OmniObject3D, yielding convincing results.

%% file: supp_tabs/ablation.tex
\begin{table}[t]
    \centering
    \small{
        \begin{tabular}{cc|c}
            \hline
            \multicolumn{2}{c|}{Methods} & \multirow{2}{*}{Fr{\'e}chet distance (FD) $\downarrow$} \\
            \cline{1-2}
            PCA & Segmentation & \\
            \hline
            \checkmark &  & 1.87 \\
            & \checkmark & 1.50 \\
            \checkmark & \checkmark & \textbf{1.43} \\
            \hline
        \end{tabular}
    }
    \caption{\textbf{Ablation Study for Data Generation.} We demonstrate the superiority of our data generation method through an ablation study. We compare the similarity between real data's OOR distributions and synthetic OOR distributions produced by our approach.}
    \label{tab:ablation}
    \vspace{-15pt}
\end{table}